\documentclass[letterpaper]{article} % DO NOT CHANGE THIS
\usepackage{aaai2026}  % DO NOT CHANGE THIS
\usepackage{times}  % DO NOT CHANGE THIS
\usepackage{helvet}  % DO NOT CHANGE THIS
\usepackage{courier}  % DO NOT CHANGE THIS
\usepackage[hyphens]{url}  % DO NOT CHANGE THIS
\usepackage{graphicx} % DO NOT CHANGE THIS
\usepackage{natbib}  % DO NOT CHANGE THIS AND DO NOT ADD ANY OPTIONS TO IT
\usepackage{caption} % DO NOT CHANGE THIS AND DO NOT ADD ANY OPTIONS TO IT
\usepackage{algorithm}
\usepackage{algorithmic}
\usepackage[utf8]{inputenc} % allow utf-8 input
\usepackage[T1]{fontenc}    % use 8-bit T1 fonts
\usepackage{url}            % simple URL typesetting
\usepackage{booktabs}       % professional-quality tables
\usepackage{amsfonts}       % blackboard math symbols
\usepackage{nicefrac}       % compact symbols for 1/2, etc.
\usepackage{microtype}      % microtypography
\usepackage{xcolor}         % colors

\usepackage{booktabs}
\usepackage{microtype}
\usepackage{epstopdf}
\usepackage{amsmath,amsfonts}
\usepackage{array}
\usepackage{enumitem}
\usepackage{placeins}
\usepackage{tabularx}
\usepackage{epstopdf}
\usepackage{mathtools}
\usepackage{caption}
\usepackage{comment}
\usepackage{multirow}
\usepackage{subcaption}
\usepackage{soul}
\usepackage{tikz}
\usetikzlibrary{tikzmark,shapes,arrows,positioning,calc,decorations.pathreplacing,shapes.geometric,decorations.pathmorphing,backgrounds,fit,matrix}
\usepackage{textcomp}

\usepackage{newfloat}
\usepackage{listings}

\usepackage{newfloat}
\usepackage{listings}
\DeclareCaptionStyle{ruled}{labelfont=normalfont,labelsep=colon,strut=off} % DO NOT CHANGE THIS
\floatstyle{ruled}
\newfloat{listing}{tb}{lst}{}
\floatname{listing}{Listing}
\title{Robust Watermarking on \\Gradient Boosting Decision Trees}
\author {
    Jun Woo Chung\textsuperscript{\rm 1},
    Yingjie Lao\textsuperscript{\rm 2},
    Weijie Zhao\textsuperscript{\rm 1}
}
\affiliations {
    \textsuperscript{\rm 1}Rochester Institute of Technology\\
    \textsuperscript{\rm 2}Tufts University\\
    jc4303@rit.edu, Yingjie.Lao@tufts.edu, wjz@cs.rit.edu
}

\begin{document}

\maketitle

\begin{abstract}
Gradient Boosting Decision Trees (GBDTs) are widely used in industry and academia for their high accuracy and efficiency, particularly on structured data. However, watermarking GBDT models remains underexplored compared to neural networks. In this work, we present the first robust watermarking framework tailored to GBDT models, utilizing in-place fine-tuning to embed imperceptible and resilient watermarks. We propose four embedding strategies, each designed to minimize impact on model accuracy while ensuring watermark robustness. Through experiments across diverse datasets, we demonstrate that our methods achieve high watermark embedding rates, low accuracy degradation, and strong resistance to post-deployment fine-tuning.
\end{abstract}


\begin{links}
     \link{AAAI Proceedings}{TBD} % Use your final arXiv ID
     \link{Code}{https://github.com/jc4303/gbdt_watermarking} % If applicable
\end{links}

\section{Introduction}

\textbf{Gradient Boosting Decision Trees.} 
Gradient Boosting Decision Trees (GBDT) have become increasingly popular within the machine learning community due to their high accuracy, interpretability, scalability, and inference speed~\cite{DBLP:conf/pakdd/FanCMKFY24, DBLP:conf/nips/IosipoiV22, DBLP:conf/nips/KeMFWCMYL17}. They often outperform neural networks, particularly when dealing with structured data containing moderate feature counts, noisy datasets, or imbalanced classes~\cite{DBLP:conf/nips/IosipoiV22, DBLP:conf/nips/McElfreshKVCRGW23}. This makes GBDT a valuable tool widely adopted across numerous applications, including privacy-sensitive and healthcare domains~\cite{DBLP:journals/corr/abs-2407-08250, Taha2025hl}.

As security considerations become increasingly critical across various domains, the widespread adoption of GBDT, as with other machine learning models, has heightened interest in securing these models as well~\cite{DBLP:conf/ccs/LawLPPSSYZZ20, DBLP:conf/uss/LuHZWH23}. Ensuring robust protection against threats such as unauthorized access, data leakage, and model tampering has become an essential area of machine learning research~\cite{DBLP:journals/csur/CinaGDVZMOBPR23, DBLP:journals/csur/RigakiG24}, and thus developing effective security measures is vital for maintaining trust and compliance in sectors heavily reliant on GBDT models. 

\textbf{Watermarking.} 
To this end, watermarking embeds identifiable patterns in a model (e.g., forcing specific predictions on selected inputs) to verify ownership and guard against tampering.~\cite{DBLP:conf/uss/AdiBCPK18, DBLP:conf/nips/GuoLWXHLL23}.

In our work, we focus on robust watermarking, which aims to embed watermarks that are resilient to further fine-tuning or other modifications~\cite{DBLP:conf/acsac/PagnottaHHPM24, DBLP:conf/uss/YanP0023}. This means that the watermark remains detectable and verifiable even with attempts to alter or erase it, thereby providing persistent evidence of provenance and safeguarding intellectual property rights~\cite{DBLP:journals/iacr/RouhaniCK18}. This is in contrast to weak watermarks, which are designed to become undetectable or degrade significantly upon modification of the model, thus clearly indicating that unauthorized alterations have been made~\cite{DBLP:conf/uss/AdiBCPK18} to a given model.

\textbf{Challenges.}
GBDT models are more complex than other tree-based models such as random forests~\cite{DBLP:conf/nips/ChenZS0BH19}, as they build trees sequentially with dependencies on prior predictions. Modifying existing trees risks cascading disruptions and degrading model accuracy.~\cite{DBLP:conf/kdd/0001LL22}. Thus, watermarking GBDT models through direct tree manipulation is a challenging endeavor. On the other hand, watermarking methods applied to neural networks (which allow subtle shifts to continuous decision boundaries) cannot be applied directly to GBDT because of the non-differentiability of tree-based models~\cite{DBLP:conf/kdd/0001LL22}.

\textbf{Approaches.} Our work addresses these issues by proposing and empirically comparing four watermark embedding techniques, each leveraging in-place updates to the initial model using strategically selected watermark samples. These techniques are: 
(a) \textit{Wrong Prediction Flip}, which embeds watermarks by tweaking samples wrongly predicted by the initial model; 
(b) \textit{Outlier Flip}, which embeds watermarks targeting outlier regions in feature space to minimize accuracy disruption; 
(c) \textit{Cluster Center Flip}, which embeds watermarks by flipping the cluster centroid prediction while trying to keep its neighboring region prediction unchanged;
(d) \textit{Confidence Flip}, which embeds watermarks by targeting correctly classified samples with the lowest prediction confidence, ensuring watermarks are embedded near decision boundaries, where model predictions are more malleable. All four of these methods are designed to embed robust watermarks while minimizing their impact on model accuracy.

\textbf{Contributions.} Our main contributions are:
\begin{itemize}[topsep=0pt, itemsep=1pt, parsep=0pt,leftmargin=*]
\item We propose robust watermarking methods for GDBT models, based on in-place fine-tuning. To the best of our knowledge, this is the first robust GDBT watermarking framework with in-place updating, and the first work to focus on robust GBDT watermarking in general.
\item We propose four watermark embedding approaches---\textit{Wrong Prediction Flip}, \textit{Outlier Flip}, \textit{Cluster Center Flip}, and \textit{Confidence Flip}---designed to minimize degradation of accuracy while embedding the robust watermarks.
\item We empirically demonstrate that our methods achieve good watermark embedding success, while incurring limited impact on the overall model accuracy. Additionally, our experiments show that the watermarks remain robust against further fine-tuning.
\end{itemize}

\section{Background and Related Work}

\subsection{Gradient Boosting}

Given a training dataset \( \mathcal{D}_{\text{train}} = \{(\mathbf{x}_i, y_i)\}_{i=1}^{N} \), where each \( \mathbf{x}_i \) is an input feature vector and \( y_i \in \{0, 1, \dots, K-1\} \) are categorical class labels, Gradient Boosted Decision Trees (GBDT) builds the predictive function \( F^{(M)}(\mathbf{x}) \) as an additive expansion of regression trees. This function can be expressed as $F^{(M)}(\mathbf{x}) = F^{(0)}(\mathbf{x}) + \sum_{m=1}^{M}\gamma_m t_m(\mathbf{x}; a_m),$ where \( F^{(0)}(\mathbf{x}) \) is an initial approximation, each \( t_m(\mathbf{x}; a_m) \) denotes a regression tree characterized by parameters \( a_m \), and \( \gamma_m \) is the scaling factor determined at each iteration. During the iterative optimization process, each new tree \( t_m(x; a_m) \) is trained to fit the pseudo-residuals, or the negative gradient of the loss function evaluated at the current approximation:
\setlength{\abovedisplayskip}{1pt}
\setlength{\belowdisplayskip}{1pt}
\begin{align}
r_{i,k}^{(m)} = -\frac{\partial L(y_i, F(\mathbf{x}_i))}{\partial F_k(\mathbf{x}_i)}\bigg|_{F_k(\mathbf{x}_i)=F_k^{(m-1)}(\mathbf{x}_i)}
\end{align}
For classification with multiple classes, the softmax function is typically employed for probability calculation: 
$p_{i,k}^{(m-1)} = \Pr(y_i = k \mid \mathbf{x}_i) = \frac{\exp(F_k^{(m-1)}(\mathbf{x}_i))}{\sum_{c=0}^{K-1}\exp(F_c^{(m-1)}(\mathbf{x}_i))},$ where \( F_k(\mathbf{x}) \) denotes the model output for class \( k \). 
Model parameters are optimized by minimizing the negative log-likelihood loss $L = -\sum_{i=1}^{N}\sum_{k=0}^{K-1} y_{i,k}\log\left(p_{i,k}^{(m-1)}\right),\quad\text{where}\quad y_{i,k} = \mathbf{1}(y_i = k)$. Training proceeds iteratively by computing the gradients and Hessians of the loss with respect to each class prediction \( F_k(x) \):
$g_{i,k}^{(m)} = -(y_{i,k} - p_{i,k}^{(m-1)})$,
$h_{i,k}^{(m)} = p_{i,k}^{(m-1)}(1 - p_{i,k}^{(m-1)})$.
Thus, each new tree \( t_m(x; a_m) \) is trained to approximate the pseudo-residuals defined explicitly as 
$r_{i,k}^{(m)} = - g_{i,k}^{(m)} = y_{i,k} - p_{i,k}^{(m-1)}.$
While this iterative fitting effectively minimizes the loss function and produces accurate and robust predictive models, the dependence of each new tree on the previous state of the model dictates that gradient boosting models are significantly more complex to modify or watermark than tree ensembles in which each tree is independent, such as random forests.

\subsection{Watermarking}

Watermarking involves embedding unique and identifiable markers within machine learning models to establish clear ownership and verify authenticity. Robust watermarking techniques are specifically designed to withstand subsequent model modifications, including fine-tuning or other adversarial alterations. Such resilience ensures that the watermark remains detectable and verifiable even after significant changes to the model. This persistent detectability enables reliable tracing of the original provenance and guards against unauthorized usage, modification, or tampering. Conversely, weaker watermarking techniques that degrade or vanish upon model alteration serve primarily to detect unauthorized modifications rather than reliably establish enduring ownership.

Watermarking has been a very active area of study for neural networks in recent years \cite{DBLP:conf/uss/AdiBCPK18, DBLP:conf/mir/UchidaNSS17}. The large data requirements, model complexity, and computing power needed to train modern neural networks have heightened interest in intellectual property protection of models. However, research on watermarking tree-based models has been far less active, since their discrete structure, limited parameter space, and lower redundancy make it more difficult to embed watermarks in a robust and unobtrusive way. One of the few works in this area by Calzavara et al.~\cite{DBLP:conf/edbt/CalzavaraCG025} targets random forests by directly modifying trees in the ensemble. However, this approach is incompatible with gradient boosting models, where trees are not independent but are sequentially constructed with dependencies based on gradients of prior trees. Zhao et al.~\cite{DBLP:conf/kdd/0001LL22} introduced a watermarking mechanism for boosted tree models, but their method focuses on fragile integrity authentication (i.e. weak watermarking) rather than robust embedding. To the best of our knowledge, our work is the first to introduce a robust watermarking framework specifically designed for gradient boosted decision trees.

\subsection{In-place Updates}

Most gradient boosting models (e.g., XGBoost) use tree addition during fine-tuning~\cite{DBLP:conf/kdd/ChenG16, DBLP:conf/nips/KeMFWCMYL17}, but this is problematic for watermarking as they can be easily removed by pruning low-contribution trees.

Thus, we implement in-place updating w.r.t. the fine-tuning process, adjusting internal parameters of existing trees rather than adding new ones. This approach integrates watermarks more deeply within the existing model, improving robustness.

Algorithm~\ref{alg:inplace} outlines our in-place update method. For each boosting iteration and each class $k$, we compute pseudo-residuals using the difference between the true gradient signals $r_{i,k}$ and the model's predicted probability $p_{i,k}$, forming a new fine-tuning dataset. For each non-terminal node in the tree, traversed in top-down depth-first order, we recompute gain scores and identify the best split. If the best new split differs from the current one, we retrain the subtree rooted at that node. Finally, we update the terminal node predictions to reflect the adjusted gradients. This procedure modifies the original model structure without expanding it.

\begin{figure}[t]
\centering
\footnotesize % <<<<<<<< Add this line to shrink text

\begin{algorithm}[H]
\small	
 \caption{GBDT In-place updating}
\label{alg:inplace}
\algsetup{linenodelimiter=.}
   \raggedright  \textbf{Input}: Initial tree ensemble $T$ (and corresponding predictive function $F$) with class labels \( y_i \in \{0, 1, \dots, K-1\} \) and $M$ iterations, fine-tuning dataset $\mathcal{D}_{\text{fine}}$\\
\raggedright \textbf{Output}: Modified tree ensemble $T'$\\
\begin{algorithmic}[1]
 
\FOR{$m = 0$ to $M - 1$}
\FOR{$k = 0$ to $K - 1$}
\STATE $\mathcal{D}_{\text{fine}}' = \{(\mathbf{x}_i,r_{i,k} - p_{i,k})\}$ for $i \in len(\mathcal{D}_{\text{fine}})$
\STATE Compute $g_{i,k}'$ and $h_{i,k}'$  w.r.t. $F_{i,k}$ and $y_{i,k}'$ as defined by $\mathcal{D}_{\text{fine}}'$
%\FOR {\textbf{each} \textit{non-terminal node} in tree $\{T_{m,k,l}\}_{l}^{L}$ where ${L}$ is the number of terminal nodes, from root to lower depth}
\FOR{\textbf{each} non-terminal node \textit{n} in tree $T_{m,k}$ (depth-first, top-down)}
    \STATE Recompute gains for $g_{i,k}'$ and $h_{i,k}'$ and best split $\textit{S'}$
    \IF{current split $\textit{S} \neq \textit{S}'$}
        \STATE Retrain subtree rooted at \textit{n}
    \ENDIF
\ENDFOR
\STATE Update prediction values at terminal nodes in $T_{m,k}$
\ENDFOR
\ENDFOR

\end{algorithmic}
\end{algorithm}
\end{figure}
\begin{figure}[htpb]

    \centering
    \includegraphics[width=\linewidth]{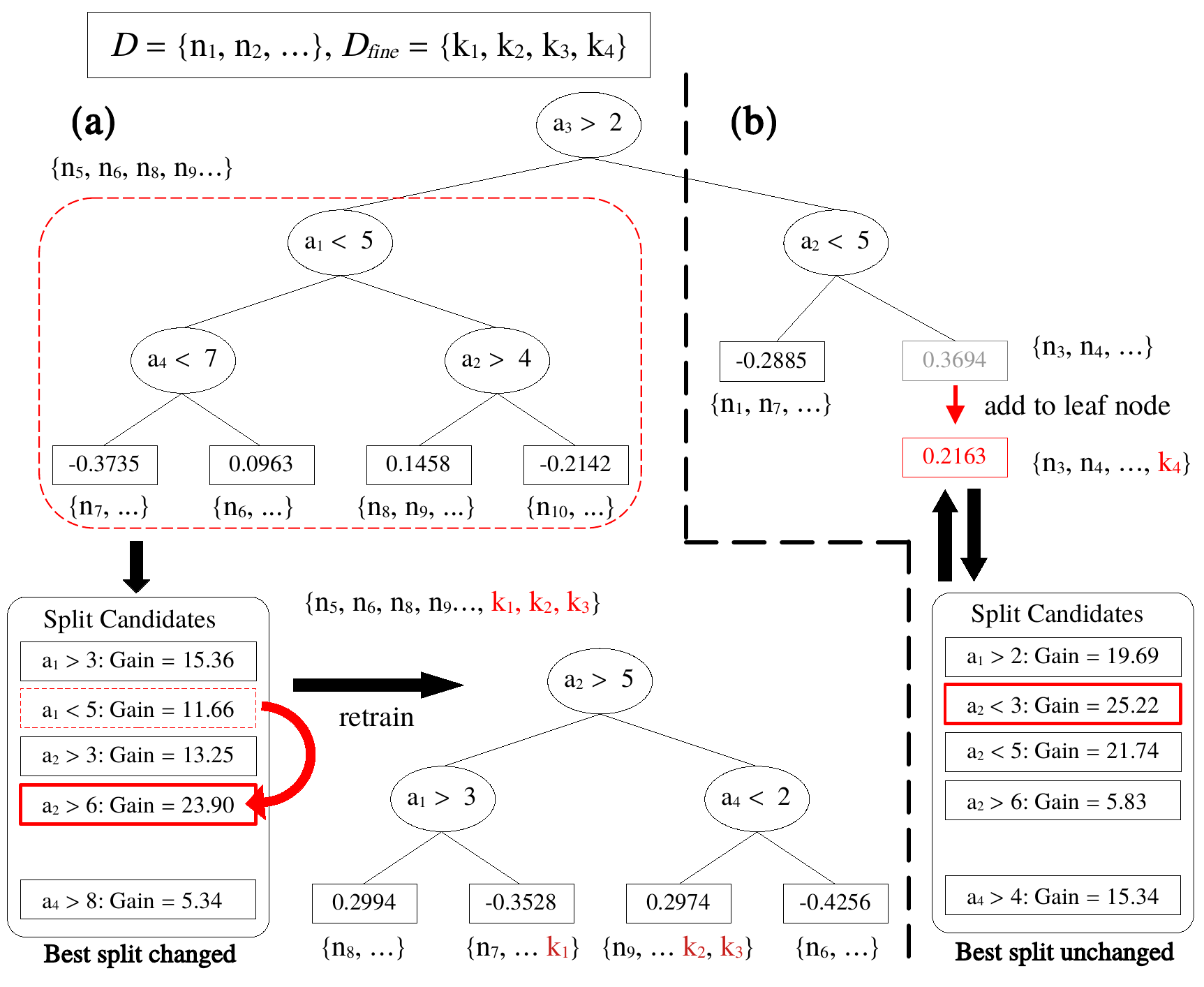} 
    \caption{Illustration of in-place updating process for a single tree for a initial gradient boosting model trained on dataset $\mathcal{D}$, as detailed in Algorithm~\ref{alg:inplace}. If the optimal split of a root node of a subtree changes due to the additional data ($\mathcal{D}_{\text{fine}}$), the corresponding subtree is retrained. If the optimal split is not changed, retraining is not needed, and only the leaf nodes for which any of the additional data corresponds to needs to be updated to reflect the update.} 
    \label{fig:inplace}
\end{figure}

% \begin{figure}[htbp]
%     \centering
%         \includegraphics[width=0.65\linewidth]{fig/inplace2.pdf} 
%     \caption{Illustration of in-place updating process for a single tree for a initial gradient boosting model trained on dataset $D$, as detailed in Algorithm~\ref{alg:inplace}. If the optimal split of a root node of a subtree changes due to the additional data ($\mathcal{D}_{\text{fine}}$), the corresponding subtree is retrained. If the optimal split is not changed, retraining is not needed, and only the leaf nodes for which any of the additional data corresponds to needs to be updated to reflect the update.} 
%     \label{fig:inplace}
% \end{figure}

\section{Embedding Watermarks in GBDT}

\subsection{Watermark Embedding Framework} \label{sec:framework}

The primary goal of our watermarking process is to enable binary information encoding within the model’s predictions. To this end, we identify a set of candidate samples \( \mathcal{C} \), from which a subset \( \mathcal{W} \subset \mathcal{C} \) of size \( k \) is selected for watermark embedding. Each sample in \( \mathcal{W} \) can represent a single bit of information: by either modifying its label (representing a '1') or retaining the original label (representing a '0').

We propose four approaches for choosing the initial watermarking candidate set \( \mathcal{C} \), as described in Section~\ref{sec:approach}. Each method relies on initial model predictions concerning a candidate dataset ($\mathcal{D}_{\text{cand}}$), which is distinct from the training or testing dataset. Initially, a model trained on a base $\mathcal{D}_{\text{train}}$ dataset is employed to generate predictions on $\mathcal{D}_{\text{cand}}$. Using these predictions, a set of \( n \) watermark candidates \( \mathcal{C} \) is identified. A subset \( \mathcal{W} \subset \mathcal{C} \) of size \( k \) is then selected for embedding, using one of the embedding selection procedures described in Section~\ref{sec:candidate_selection}.

Watermarks are embedded by fine-tuning the model using a dataset constructed from the selected watermark samples, where the ground truth label for each sample is modified. In general, the new label is set to the most confident incorrect prediction, excluding both the ground truth and the model's original prediction (the two of which will only be different for the \textit{Wrong Prediction Flip} method):
\setlength{\abovedisplayskip}{1pt}
\setlength{\belowdisplayskip}{1pt}
\begin{align}
y_i^{\text{wm}} = \underset{c \ne y_i, \; c \ne \hat{y}_i}{\text{argmax}} \; F_c(\mathbf{x}_i) \; | \; \hat{y}_i = \arg\max F(\mathbf{x}_i)
\label{eq:wm}
\end{align}

The exception is the \textit{Cluster Center Flip} approach, which introduces additional constraints, as detailed in Section~\ref{sec:approach}.

\subsection{Candidate Dataset} \label{sec:candidate_dataset}

Watermark candidates are selected from either the training data or a separate dataset (e.g., a split or independent source), denoted \( \mathcal{D}_{\text{cand}} \). This may be the training set \( \mathcal{D}_{\text{train}} \) or a separate holdout set \( \mathcal{D}_{\text{holdout}} \). Using a separate dataset avoids interference during watermark fine-tuning from gradients related to unmodified versions of the same samples in the training set. It reflects scenarios where watermarking is applied post hoc by third parties (e.g., model purchasers). 

In contrast, selecting candidates from training data avoids new data but introduces conflicts---as the model has seen the unmodified samples, fine-tuning on modified versions may not shift decision boundaries enough. We resolve this by duplicating the samples by a factor ${d}_{cand=train}$ in fine-tuning to ensure they dominate the relevant gradient updates.

We note that it is not guaranteed that the gradient w.r.t. the watermark sample is modified in the correct direction; for the application of a watermarked sample $\mathbf{x}_i$ with label \( y_i^{\text{wm}} \neq y_i \) with multiplier $r$, $g_{i,k}^{\text{wm}} = -(y_{i,k}^{\text{wm}} - p_{i,k})$ and thus $g_{i,k}^{\text{total}} = - (y_{i,k}^{\text{wm}} - p_{i,k}) r - (y_{i,k} - p_{i,k})$. Accordingly, the gradient for the class $k_{y_{i}} = y_{i}$ corresponding to the original $y_i$ is $g_{y_{i}} = -1 + (1 + r)p_{i,{y_{i}}}$ (which is positive only if $p_{i,{y_{i}}} > 1/(1+r)$) and the gradient for the watermark class $k_{y_{i}^{\text{wm}}} = y_{i}^{\text{wm}}$ is $g_{y_{i}^{\text{wm}}} = -r + (1 + r)p_{i,{y_{i}^{\text{wm}}}}$ (which is negative only if $p_{i,{y_{i}^{\text{wm}}}} < r/(1+r)$). Thus, for any arbitrary $r$, we cannot strictly guarantee that the gradient for $\mathbf{x}_i$ is negative for the watermark class $k_{y_{i}^{\text{wm}}}$.

\subsection{Approaches} \label{sec:approach}
%\hl{Include illustrative figures demonstrating each embedding method clearly.}

\begin{figure}[tb]
    \centering
        \includegraphics[width=\linewidth]{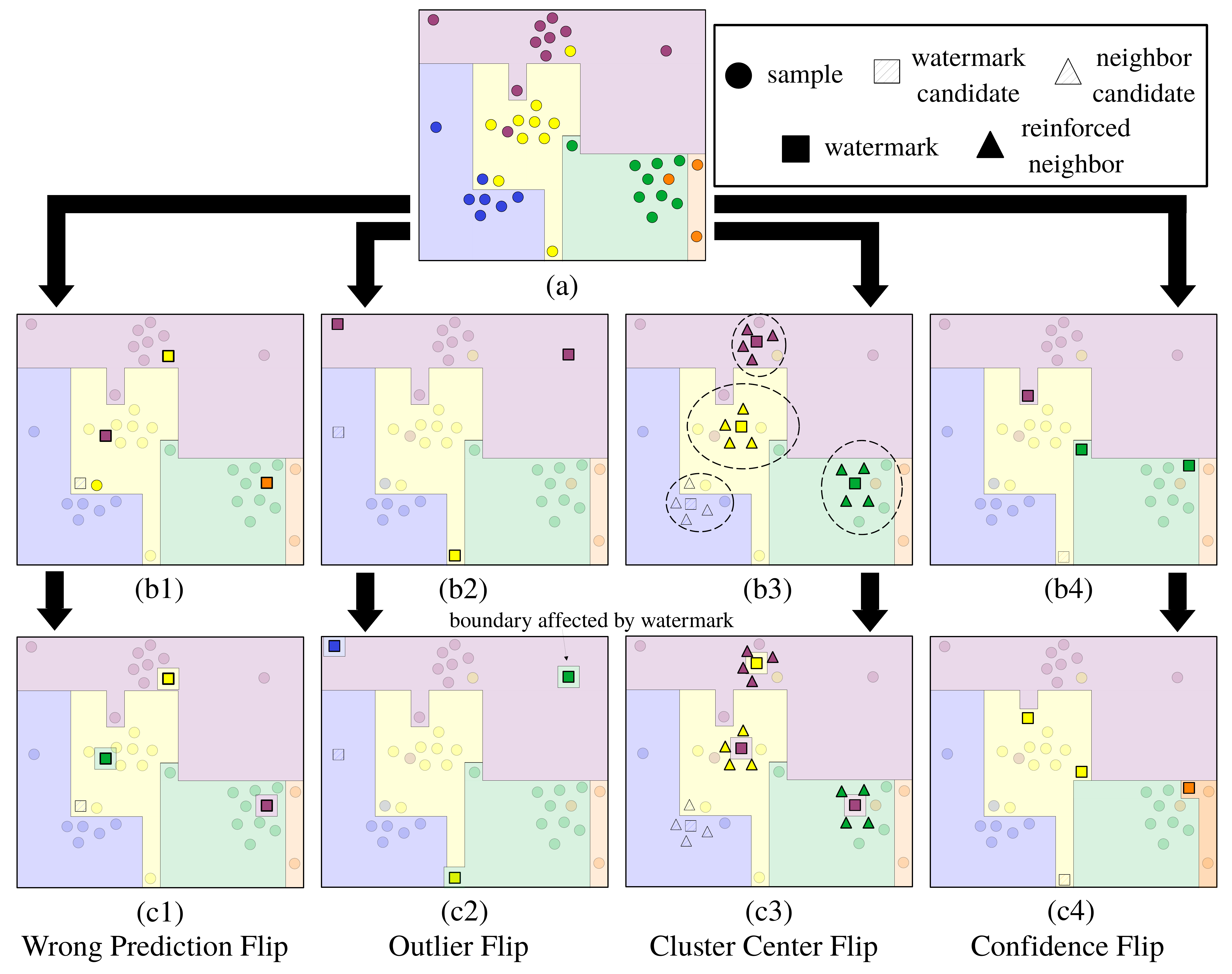} 
    \caption{Example of an initial model, watermark selection, and embedding for the different selection methods. In (a), the decision boundaries and predictions of the initial model are shown as background shading, with the $\mathcal{D}_{\text{cand}}$ dataset overlaid in feature space using colored circles to represent ground truth labels. (b1) highlights watermark candidates for \textit{the Wrong Prediction Flip} approach, or samples misclassified by the model, evident where circle colors differ from the background, and the selected watermarks outlined with thick edges for $\textit{n} = 4$ and $\textit{k} = 3$. (c1) displays the modified labels used during fine-tuning (second most probable incorrect class), and the anticipated boundary adjustments. (b2) and (c2) follow the same visual conventions for the \textit{Outlier Flip} approach, where selected watermarks are the samples most distant from others in feature space, and the new label is the highest-probability incorrect class. (b3) shows the \textit{Cluster Center Flip} strategy, where cluster centroids are selected as watermark candidates. Their $\textit{l}$ nearest neighbors, which reinforce the original labels, are marked with triangles. (c3) depicts the label and boundary changes resulting from tuning the watermarks and neighbors. Finally, (b4) and (c4) illustrate the \textit{Confidence Flip} approach, where low-confidence correct predictions (often near decision boundaries) are selected.}
    \label{fig:all_method}
\end{figure}

We decribe the approaches below, and demonstrate them visually in Figure~\ref{fig:all_method}.

\textbf{Wrong Prediction Flip}. In this approach, watermark candidates are drawn from the samples in the $\mathcal{D}_{\text{cand}}$ dataset that the model initially classifies incorrectly (i.e., the predicted label differs from the ground truth). From these samples, \textit{n} samples (which have the lowest confidence) are chosen as watermark candidates:
{\scriptsize
\begin{align}
\bigl\{\mathcal{C} = \underset{\mathbf{x}_i \in \mathcal{D}}{\text{argmin}_n} \; F_{\hat{y}_i}(\mathbf{x}_i)\bigr\} 
\quad \text{where} \quad 
\mathcal{D} = \bigl\{ \mathbf{x}_i \bigm| y_i \ne \hat{y}_i, \; (\mathbf{x}_i, y_i) \in \mathcal{D}_{\text{cand}} \bigr\}
\label{eq:wrong}
\end{align}
}

\begin{figure}
    \centering
        \includegraphics[width=0.5\linewidth]{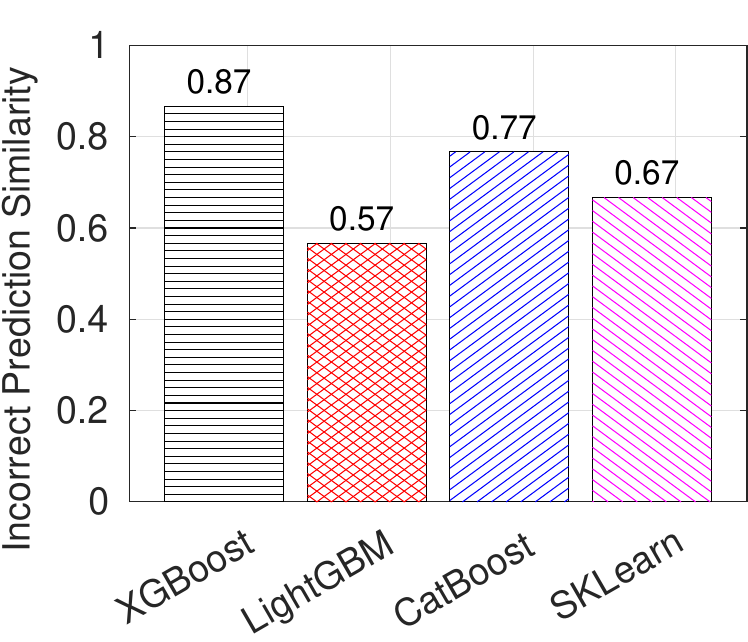} 
    \caption{The proportion of incorrect predictions made by our initial model that are also incorrect w.r.t. models trained using other GBDT libraries w.r.t. \textit{optdigit} dataset. The relatively high similarity demonstrates that simply using the incorrect predictions as watermarks risks them simply being ``hard'' samples, thus meaning unrelated models can make similar predictions to the watermark,  leading to ambiguity problems.} 
    \label{fig:sim_other}
\end{figure}

where $\hat{y}_i$ is the initial model prediction, defined by Equation~\ref{eq:wm}. The modified ground truth for watermark embedding is set to the second most probable incorrect prediction rather than the initially predicted label, which often reflect difficult cases misclassified even by unrelated models (Figure~\ref{fig:sim_other}).

This aims to minimize the impact on general model accuracy, as embedding occurs in regions already prone to misclassification, subject to availability of watermark candidates \textit{n} which are inherently dependent on the dataset. Additionally, for cases where $\mathcal{D}_{\text{cand}} = \mathcal{D}_{\text{train}}$, the tendency of GDBT to be very accurate w.r.t. training data means that substantial numbers of wrongly predicted samples are unlikely.

\textbf{Outlier Flip}. This method selects \textit{n} samples correctly predicted by the initial model yet furthest in feature space from other elements within the $\mathcal{D}_{\text{cand}}$ dataset. The definition of ``furthest'' can be chosen w.r.t. context; in our experiments, we apply the \textit{k}-Means algorithm, with the cluster count $m$ that maximizes the silhouette score. Afterwards, we select the $n$ samples that are furthest from any cluster centroid $ \boldsymbol{\mu}_j \mid j \in \{1, 2, \dots, m\}$ as the watermark candidates:

{\scriptsize
\begin{align}
\mathcal{C} = \bigl\{\underset{\mathbf{x}_i \in \mathcal{D}}{\text{argmax}_n} \; \min_{j \in \{1, \dots, m\}} \lVert \mathbf{x}_i - \boldsymbol{\mu}_j \rVert\bigr\} \notag\\
\text{where} \quad \mathcal{D} = \bigl\{ \mathbf{x}_i \bigm| y_i = \hat{y}_i,  \; (\mathbf{x}_i, y_i) \in \mathcal{D}_{\text{cand}} \bigr\}
\label{eq:out}
\end{align}
}
From these candidates, \textit{k} watermarks are selected using the criteria described in Section~\ref{sec:candidate_selection}.

The primary objective of this method is to embed watermarks in sparse (w.r.t. gneneral population) regions in feature space, limiting accuracy impact while enhancing robustness against conventional fine-tuning, presuming similar distributions between fine-tuning datasets and the $\mathcal{D}_{\text{cand}}$ dataset. However, this method's accuracy preservation and resiliency against further fine-tuning depends on this distributional similarity, which may not always hold.

\textbf{Cluster Center Flip}. In this method, \textit{n} clusters are identified within the watermark dataset using clustering algorithms (e.g., \textit{k}-Means or DBSCAN~\cite{DBLP:conf/kdd/EsterKSX96}). For each cluster, the element closest to the centroid $ \boldsymbol{\mu}_j \mid j \in \{1, 2, \dots, n\}$ ($\mathcal{C}$) and its \textit{l} nearest neighbors ($\mathcal{C'}$) are selected:

{\scriptsize
\begin{align}
\mathcal{C} = \bigl\{ \underset{\mathbf{x}_i \in \mathcal{D}}{\text{argmin}} \; \lVert \mathbf{x}_i - \boldsymbol{\mu}_j \rVert \; \bigm| \; j = 1, \dots, n \bigr\}, \notag\\
\mathcal{C'} = \bigl\{ \underset{\mathbf{x}_i \in \mathcal{D}, \;  \mathbf{x}_i \notin \mathcal{C}}{\text{argmin}_l} \; \lVert \mathbf{x}_i - \boldsymbol{\mathbf{x}}_j \rVert \; \bigm| \; \mathbf{x}_j \in \mathcal{C} \bigr\}
\label{eq:near}
\end{align}
}
From $\mathcal{C}$ and $\mathcal{C'}$, $k$ watermarks are selected following the processes described in Section~\ref{sec:candidate_selection}, and their corresponding $kl$ neighbors are also added as the embedding dataset. Each cluster centroid ($w \in \mathcal{C}$) would be embedded as a watermark by assigning it the highest-probability incorrect classification, while the neighboring samples ($w' \in \mathcal{C'}$) would retain their original, correct labels.

This creates localized disruptions (``holes'') in the decision boundary by anchoring the watermark with correctly labeled surrounding points, thereby minimizing any broader impact on general accuracy independent of dataset distributions. However, reinforcing neighbors to maintain correct predictions could potentially dilute the intended watermark signal due the opposing pressure caused by the neighbors of the watermark. To counteract this, we duplicate the centroid group $\mathcal{C}$ once in the watermarking dataset.

\textbf{Confidence Flip}. This method selects watermark candidates from among the correctly predicted samples in the $\mathcal{D}_{\text{cand}}$ dataset that have the lowest confidence, i.e. it finds a watermark candidate set $C$ which satisfies 
\setlength{\abovedisplayskip}{1pt}
\setlength{\belowdisplayskip}{1pt}
{\scriptsize
\begin{align}
\mathcal{C} = \underset{\mathbf{x}_i \in \mathcal{D}}{\text{argmin}_n} \; F_{y_i}(\mathbf{x}_i) \quad \text{where} \quad \mathcal{D} = \left\{ \mathbf{x}_i \mid y_i = \hat{y}_i, \; (\mathbf{x}_i, y_i) \in \mathcal{D}_{\text{cand}}  \right\}
\label{eq:conf}
\end{align}}
 given an initial trained predictive function $F(\mathbf{x})$. Such elements are likely to lie near class boundaries and can thus be shifted while minimizing collateral effects on confidently classified regions and thus with robust, high-confidence predictions while maximizing susceptibility to the embedding.

\subsection{Candidate Selection} \label{sec:candidate_selection}

All three watermark methods described in Section~\ref{sec:approach} first define a set of watermark candidates $\mathcal{C}$. From these, a subset of \textit{k} watermarks $\mathcal{W}$ is subsequently selected for embedding. We describe two approaches w.r.t. this selection process;

\begin{table*}[bhtp]
\centering
\scriptsize
\renewcommand{\arraystretch}{0.60}
\caption{Watermarking effectiveness for $\mathcal{D}_{\text{cand}} = \mathcal{D}_{\text{train}}$ and $\textit{d}_{cand=train} = 5$. For brevity, method names are abbreviated as Cluster, Outlier, Wrong, Conf. (\textit{Confidence Flip}), and Random (random selection baseline) with selection strategies noted in parentheses: (Conf) (lowest-confidence) and (Dist) (maximum-distance). This is repeated for all further results. Success rate is generally high for all methods, with the \textit{Random Flip} baseline being lower than the other methods by a significant gap.}
\label{tab:watermark_trainonly_dup5_nonsep1_near2}
\begin{tabular}{@{\hskip 3pt}l@{\hskip 3pt}@{\hskip 3pt}c@{\hskip 3pt}c@{\hskip 3pt}c@{\hskip 3pt}c@{\hskip 3pt}c@{\hskip 3pt}c@{\hskip 3pt}c@{\hskip 3pt}c@{\hskip 3pt}c@{\hskip 3pt}c@{\hskip 3pt}c@{\hskip 3pt}c@{\hskip 3pt}c@{\hskip 3pt}c@{\hskip 3pt}c@{\hskip 3pt}c@{\hskip 3pt}c@{\hskip 3pt}c@{\hskip 3pt}|@{\hskip 3pt}c@{\hskip 3pt}c@{\hskip 3pt}c@{}}
\toprule
Dataset & \multicolumn{3}{@{\hskip -0.2pt}c}{Avila} & \multicolumn{3}{@{\hskip -0.2pt}c}{Img Seg.} & \multicolumn{3}{@{\hskip -0.2pt}c}{Letter Recognition} & \multicolumn{3}{@{\hskip -0.2pt}c}{optdigits} & \multicolumn{3}{@{\hskip -0.2pt}c}{pendigits} & \multicolumn{3}{@{\hskip -0.2pt}c}{Wine Quality} & \multicolumn{3}{@{\hskip -0.2pt}c}{average} \\
\cmidrule(l{1pt}r{6pt}){2-4} \cmidrule(l{1pt}r{6pt}){5-7} \cmidrule(l{1pt}r{6pt}){8-10} \cmidrule(l{1pt}r{6pt}){11-13} \cmidrule(l{1pt}r{6pt}){14-16} \cmidrule(l{1pt}r{6pt}){17-19} \cmidrule(l{1pt}r{6pt}){20-22}
Method | Ratio & 0.001 & 0.01 & 0.1 & 0.001 & 0.01 & 0.1 & 0.001 & 0.01 & 0.1 & 0.001 & 0.01 & 0.1 & 0.001 & 0.01 & 0.1 & 0.001 & 0.01 & 0.1 & 0.001 & 0.01 & 0.1 \\
\midrule
Cluster (Conf) & \textbf{1.000} & \textbf{1.000} & \textbf{1.000} & \textbf{1.000} & \textbf{1.000} & \textbf{1.000} & \textbf{1.000} & 0.880 & 0.993 & 0.500 & \textbf{1.000} & \textbf{1.000} & 0.250 & \textbf{1.000} & \textbf{1.000} & \textbf{1.000} & \textbf{1.000} & \textbf{1.000} & 0.792 & 0.980 & 0.999 \\
Cluster (Dist) & \textbf{1.000} & \textbf{1.000} & \textbf{1.000} & \textbf{1.000} & \textbf{1.000} & \textbf{1.000} & \textbf{1.000} & 0.507 & 0.968 & 0.500 & \textbf{1.000} & \textbf{1.000} & 0.750 & \textbf{1.000} & \textbf{1.000} & \textbf{1.000} & \textbf{1.000} & \textbf{1.000} & 0.875 & 0.918 & 0.995 \\
Outlier (Conf) & \textbf{1.000} & \textbf{1.000} & \textbf{1.000} & \textbf{1.000} & \textbf{1.000} & \textbf{1.000} & 0.875 & 0.720 & 0.996 & 0.500 & \textbf{1.000} & \textbf{1.000} & \textbf{1.000} & \textbf{1.000} & \textbf{1.000} & \textbf{1.000} & \textbf{1.000} & \textbf{1.000} & 0.896 & 0.953 & 0.999 \\
Outlier (Dist) & \textbf{1.000} & \textbf{1.000} & \textbf{1.000} & \textbf{1.000} & \textbf{1.000} & \textbf{1.000} & \textbf{1.000} & 0.760 & \textbf{1.000} & \textbf{1.000} & \textbf{1.000} & \textbf{1.000} & 0.250 & \textbf{1.000} & \textbf{1.000} & \textbf{1.000} & 0.960 & \textbf{1.000} & 0.875 & 0.953 & \textbf{1.000} \\
Wrong (Conf) & --- & --- & --- & --- & --- & --- & --- & --- & --- & --- & --- & --- & --- & --- & --- & \textbf{1.000} & \textbf{1.000} & --- & \textbf{1.000} & \textbf{1.000} & --- \\
Wrong (Dist) & --- & --- & --- & --- & --- & --- & --- & --- & --- & --- & --- & --- & --- & --- & --- & \textbf{1.000} & \textbf{1.000} & --- & \textbf{1.000} & \textbf{1.000} & --- \\
Conf. (Conf) & \textbf{1.000} & 0.962 & \textbf{1.000} & \textbf{1.000} & \textbf{1.000} & \textbf{1.000} & 0.875 & \textbf{0.920} & 0.992 & 0.500 & 0.850 & \textbf{1.000} & 0.250 & 0.974 & \textbf{1.000} & \textbf{1.000} & \textbf{1.000} & \textbf{1.000} & 0.771 & 0.951 & 0.999 \\
Conf. (Dist) & \textbf{1.000} & \textbf{1.000} & \textbf{1.000} & \textbf{1.000} & \textbf{1.000} & \textbf{1.000} & 0.625 & 0.907 & 0.996 & 0.500 & 0.850 & \textbf{1.000} & 0.750 & \textbf{1.000} & \textbf{1.000} & \textbf{1.000} & \textbf{1.000} & \textbf{1.000} & 0.812 & 0.959 & 0.999 \\
Random (Conf) & 0.667 & 0.830 & 0.975 & \textbf{1.000} & \textbf{1.000} & \textbf{1.000} & \textbf{1.000} & 0.480 & 0.917 & 0.000 & 0.750 & \textbf{1.000} & 0.500 & 0.895 & \textbf{1.000} & \textbf{1.000} & 0.960 & \textbf{1.000} & 0.694 & 0.819 & 0.982 \\
Random (Dist) & \textbf{1.000} & \textbf{1.000} & 0.996 & \textbf{1.000} & \textbf{1.000} & \textbf{1.000} & \textbf{1.000} & 0.507 & 0.953 & 0.000 & 0.800 & \textbf{1.000} & 0.250 & 0.974 & \textbf{1.000} & \textbf{1.000} & \textbf{1.000} & 0.996 & 0.708 & 0.880 & 0.991 \\
\bottomrule
\end{tabular}

\end{table*}
\textbf{Lowest Confidence}. To identify low-confidence predictions, we rank the samples in \( \mathcal{C} \) w.r.t. prediction confidence and select the lowest values. Specifically, we construct the watermark set \( \mathcal{W} \subset \mathcal{C} \) of size \( k \) by solving
$\mathcal{W} = \underset{\substack{\mathcal{S} \subset \mathcal{C}, \textit{ } |\mathcal{S}| = k}}{\arg\min} \; \sum_{\mathbf{x}_i \in \mathcal{S}} F_{\hat{y}_i}(\mathbf{x}_i)$.
By embedding watermarks into these low-confidence samples, we aim to reduce potential accuracy degradation, as these samples already lie near class boundaries. This makes their decision assignments easier to shift and less likely to affect confident regions.

\textbf{Maximum Distance}. Candidates are chosen to maximize distance from all other watermarks. This criterion aims to minimize unintended interactions between watermarks by spatially isolating them. Embedding watermarks in distant and sparsely populated regions mitigates the risk of collateral decision boundary shifts affecting nearby candidates. 

This corresponds to the maximum diversity problem~\cite{DBLP:journals/eor/LozanoMG11}, which is well-known to be NP-hard but also that an optimal answer to this problem is guaranteed to be less than twice as efficient as a greedy method that simply selects the furthest point from all the selected samples~\cite{DBLP:journals/tcs/Gonzalez85}. To that end, we apply this substantially faster and simpler solution, as this selection process serves primarily as a supplementary safeguard rather than a pivotal error reduction tool.
%-Aim to minimize

\section{Experiments} \label{sec:exp}

We evaluate the effectiveness of our method through experiments conducted on a range of public datasets that are standard--- Avila~\cite{avila459}, Image Segmentation (or Img Seg.)~\cite{imagesegmentation50}, Letter Recognition~\cite{letterrecognition59}, Wine Quality~\cite{DBLP:journals/dss/CortezCAMR09}, Optical Recognition of Handwritten Digits (or optdigits)~\cite{DBLP:journals/tsmc/XuKS92}, and Pen-Based Recognition of Handwritten Digits
(or pendigits)~\cite{pendigits81}. As detailed previously, there has been limited prior work specifically addressing watermarking for gradient boosting decision trees (GBDTs). Thus, we compare our approach against an intuitive baseline: randomly selected watermark candidates drawn from $\mathcal{D}_{\text{cand}}$. The watermarks $\mathcal{W}$ are chosen using the selection strategies described in Section~\ref{sec:candidate_selection}.

We test our method under two distinct scenarios, also described in Section~\ref{sec:framework}. In the first, watermark samples are drawn from the same dataset used for training, i.e., $\mathcal{D}_{\text{cand}} = \mathcal{D}_{\text{train}}$. This setting reflects cases where watermarking is performed internally. In the second, watermarking is performed using a dataset disjoint from the training set. This simulates scenarios where watermarking is applied post hoc—e.g., for outsourced or externally maintained models. In this case, the full dataset is partitioned into four disjoint subsets: $\mathcal{D}_{\text{train}}$, $\mathcal{D}_{\text{cand}}$, $\mathcal{D}_{\text{test}}$, and $\mathcal{D}_{\text{fine}}$. The training set is split 80:20 to form $\mathcal{D}_{\text{train}}$ and $\mathcal{D}_{\text{cand}}$. For both cases, the test set is split 80:20 to yield $\mathcal{D}_{\text{test}}$ and $\mathcal{D}_{\text{fine}}$, which is used in fine-tuning w.r.t. Section~\ref{sec:ft_robust}. If a dataset provides distinct training and test sets, we use them directly. Otherwise, we partition the dataset using an 80:20 train-test split. As detailed in Section~\ref{sec:candidate_dataset}, for $\mathcal{D}_{\text{cand}} = \mathcal{D}_{\text{train}}$, $\mathcal{W}$ is duplicated. We set this duplication factor $\textit{d}_{cand=train} = 5$. 

We assume all watermark bits in $\mathcal{W}$ are set to 1, providing a stress test for effectiveness and robustness. We evaluate watermark ratios ($|\mathcal{W}|/|\mathcal{D}_{train}|$) of 0.001, 0.01, and 0.1 across scenarios, sampling strategies (Section~\ref{sec:approach}), and selection strategies (Section~\ref{sec:candidate_selection}). Unless otherwise specified, all models are trained for 200 iterations during all phases, and we set $|\mathcal{C}| = 2|\mathcal{W}|$. Below, we describe the characteristics under consideration, the experimental setup, and results.

\subsection{Watermarking Effectiveness} \label{sec:w_eff}
As we do not manipulate the model itself directly to insert the watermark, we do not have an innate guarantee that the watermarks are correctly embedded post training. To show that our method can embed watermarks with reasonable reliability, we apply the following process:
An initial model is trained on the $\mathcal{D}_{\text{train}}$ dataset, which in turn is used for prediction on the on the $\mathcal{D}_{\text{cand}}$ dataset to identify $n$ candidate samples adhering to the requirements detailed in Section~\ref{sec:approach}. A subset of size \textit{k} of these candidates is selected and modified following the approaches detailed in Section~\ref{sec:candidate_selection}. The initial model is then watermarked (i.e. fine-tuned) using these samples---i.e. with the dataset $\mathcal{W}' = \left\{ c_i' = (\mathbf{x}_i, y_i^{\text{wm}}) \; \middle| \; (\mathbf{x}_i, y_i) \in \mathcal{W}, \; y_i^{\text{wm}} \ne y_i \right\}$, where $y_i^{\text{wm}}$ is as defined by Equation~\ref{eq:wm}.

Watermarking effectiveness, denoted $\mathcal{A}_{\text{wm}}$, is measured by the proportion of watermark samples identified as the watermark label by the watermarked model $F^{\text{wm}}$:
\setlength{\abovedisplayskip}{1pt}
\setlength{\belowdisplayskip}{1pt}
\begin{align}
\mathcal{A}_{\text{wm}} = \frac{1}{|\mathcal{W}'|} \sum_{(\mathbf{x}_i, y_i^{\text{wm}}) \in \mathcal{W}'} \mathbf{1} (F^{\text{wm}}(\mathbf{x}_i) = y_i^{\text{wm}})
\label{eq:wat}
\end{align}
This effectiveness can be increased arbitrarily by duplicating $\mathcal{W}'$ in the fine-tuning dataset during watermarking, but at the cost of other issues such as model accuracy. We do not apply this in all further testing, to prevent overfitting to watermark samples obscuring relative performance.

The results for $\mathcal{D}_{\text{cand}} = \mathcal{D}_{\text{train}}$ are presented in Table~\ref{tab:watermark_trainonly_dup5_nonsep1_near2},and those where $\mathcal{D}_{\text{cand}} \neq \mathcal{D}_{\text{train}}$ in Table~\ref{tab:watermark_sepcand_dup5_nonsep1_near2}. In both cases, general success rates are good; \textit{Wrong Prediction Flip} also has very high success rates where it is applicable, but the aspect that it requires samples that yield wrong predictions significantly limits the number of watermarks that are applicable, especially for $\mathcal{D}_{\text{cand}} = \mathcal{D}_{\text{train}}$, as outlined in Section~\ref{sec:approach}.

\subsection{General Accuracy} \label{sec:gen_acc}

Maintaining the general predictive performance of the model after watermark embedding is critical, as the practical utility of a watermarking scheme depends heavily on minimizing accuracy degradation for non-watermarked data. We evaluate the general accuracy, \( \mathcal{A}_{\text{model}} \), by assessing the performance of the fine-tuned model on the standard test set \( \mathcal{D}_{\text{test}} \):
\setlength{\abovedisplayskip}{1pt}
\setlength{\belowdisplayskip}{1pt}
{
\begin{align}
\mathcal{A}_{\text{model}} = \frac{1}{|\mathcal{D}_{\text{test}}|} \sum_{(\mathbf{x}_i, y_i) \in \mathcal{D}_{\text{test}}} \mathbf{1} \left( F^{\text{wm}}(\mathbf{x}_i) = y_i \right)
\label{eq:gen_acc}
\end{align}}
We emphasize that \( \mathcal{A}_{\text{model}} \) alone does not capture practical watermarking effectiveness. Intuitively, methods with lower \( \mathcal{A}_{\text{wm}} \) would induce smaller perturbations to the original model and consequently preserve higher \( \mathcal{A}_{\text{model}} \); however, this reduced deviation is at the expense of weaker watermark  embedding, limiting utility in practice.  Thus, we define the \emph{adjusted model accuracy} as \( \mathcal{A}_{\text{model}}' = \mathcal{A}_{\text{model}} \cdot \mathcal{A}_{\text{wm}} \), which jointly considers generalization and watermark strength to provide a more comprehensive evaluation. Table~\ref{tab:adj_model_accuracy_trainonly_dup5_nonsep1_near2} presents results for the case where \( \mathcal{D}_{\text{cand}} = \mathcal{D}_{\text{train}} \), while Table~\ref{tab:adj_model_accuracy_sepcand_dup5_nonsep1_near2} reports on the setting with $\mathcal{D}_{\text{cand}} \neq \mathcal{D}_{\text{train}}$.

For $\mathcal{D}_{\text{cand}} \neq \mathcal{D}_{\text{train}}$ we observe that \textit{Cluster Flip} and \textit{Wrong Prediction Flip} generally yield the highest adjusted model accuracy, provided the latter has access to a sufficient number of incorrect predictions. These methods are explicitly designed to minimize their impact on generalization: \textit{Cluster Flip} anchors watermarks near existing decision boundaries using nearest-neighbor heuristics, while \textit{Wrong Prediction Flip} restricts modifications to already misclassified samples. In contrast, \textit{Outlier Flip} and \textit{Confidence Flip} only apply the implicit expectation that that outliers or low-confidence samples will generally occupy sparsely populated regions of the feature space for this purpose. For $\mathcal{D}_{\text{cand}} = \mathcal{D}_{\text{train}}$ this is less visible, although a gap between the methods and baseline is still visible especially for $|\mathcal{W}|/|\mathcal{D}_{train}| = 0.01 \text{ and } 0.1$.

% \begin{figure*}[htbp]
%     \centering
%     \subfloat[Subfigure 1 caption]{\includegraphics[width=0.24\textwidth]{fig/wine_quality_random_False_sepCand_True_model_accuracy_plot.eps}}\hfill
%     \subfloat[Subfigure 2 caption]{\includegraphics[width=0.24\textwidth]{fig/image_segmentation_random_False_sepCand_True_model_accuracy_plot.eps}}\hfill
%      \subfloat[Subfigure 3 caption]{\includegraphics[width=0.24\textwidth]{fig/pendigits_random_False_sepCand_True_model_accuracy_plot.eps}}\hfill
%     \subfloat[Subfigure 4 caption]{\includegraphics[width=0.24\textwidth]{fig/optdigits_random_False_sepCand_True_model_accuracy_plot.eps}}\hfill
%      \subfloat[Subfigure 5 caption]{\includegraphics[width=0.24\textwidth]{fig/wine_quality_random_False_sepCand_True_model_accuracy_plot.eps}}\hfill
%         \subfloat[Subfigure 6 caption]{\includegraphics[width=0.24\textwidth]{fig/wine_quality_random_False_sepCand_True_model_accuracy_plot.eps}}\hfill
%        \subfloat[Subfigure 7 caption]{\includegraphics[width=0.24\textwidth]{fig/wine_quality_random_False_sepCand_True_model_accuracy_plot.eps}}\hfill
%       \subfloat[Subfigure 8 caption]{\includegraphics[width=0.24\textwidth]{fig/wine_quality_random_False_sepCand_True_model_accuracy_plot.eps}}\hfill
      
%     \caption{General accuracy w.r.t. test dataset of the watermarked models, compared to the pre-watermarking accuracy. }
%     \label{fig:general_accuracy}
% \end{figure*}

\subsection{Fine-tuning Robustness} \label{sec:ft_robust}
A central concern of our work is watermark robustness to further fine-tuning. To evaluate this, we fine-tune the  watermarked model $F^{\text{wm}}$ using the $\mathcal{D}_{\text{fine}}$ dataset. This simulates scenarios where the model is modified after deployment. We define robustness as the proportion of correctly embedded watermark samples that remain intact (i.e., still trigger the intended model response) after such tuning, i.e. for the further fine-tuned watermarked model $F^{\text{wm}'}$:
\setlength{\abovedisplayskip}{1pt}
\setlength{\belowdisplayskip}{1pt}
{\scriptsize
\begin{align}
\text{Fine-tuning robustness} = \notag\\ \frac{
    \sum_{(\mathbf{x}_i, y_i^{\text{wm}}) \in \mathcal{W}'} \mathbf{1}\left(F^{\text{wm}}(\mathbf{x}_i) = y_i^{\text{wm}} \wedge F^{\text{wm}'}(\mathbf{x}_i) = y_i^{\text{wm}}\right)
}{
    \sum_{(\mathbf{x}_i, y_i^{\text{wm}}) \in \mathcal{W}'} \mathbf{1}\left(F^{\text{wm}}(\mathbf{x}_i) = y_i^{\text{wm}}\right)
}
\label{eq:ft_robust}
\end{align}}
As Eq.~\ref{eq:ft_robust} considers only successful watermark predictions, we do not adjust for effectiveness again. Table~\ref{tab:adj_robustness_trainonly_dup5_nonsep1_near2} reports results where \( \mathcal{D}_{\text{cand}} = \mathcal{D}_{\text{train}} \), while Table~\ref{tab:adj_robustness_sepcand_dup5_nonsep1_near2} shows outcomes where \( \mathcal{D}_{\text{cand}} \neq \mathcal{D}_{\text{train}} \). The proposed methods generally show good robustness versus the baseline.

%Generally, white-box attacks against watermarks are extremely difficult to attack against for most ML models~\cite{??} and gradient boosting models (or other tree ensemble models) are by architectural nature 

\subsection{Selection Strategy}
Based on our results in Section~\ref{sec:exp}, we synthesize our findings to strategic guidance in relation to the contexts and proposed mechanisms by which each embedding strategy performs best, as well as our recommendations in Table~\ref{tab:selection-guidance}.

\section{Conclusion}
We propose the first robust watermarking framework for GBDT models, addressing their sequential and non-differentiable structure. Using in-place fine-tuning and four embedding strategies, we show that resilient watermarks can be embedded with minimal impact on performance. Our methods achieve high success rates, preserve accuracy, and remain robust to fine-tuning.

\section*{Acknowledgments}
This work is partially supported by the National Science Foundation award 2247619.

\begin{table*}[bhtp]

\centering
\scriptsize
\caption{Watermarking effectiveness for $\mathcal{D}_{\text{cand}} \neq \mathcal{D}_{\text{train}}$, without $\mathcal{W}'$ duplication. On average  \textit{Confidence Flip} has the highest success rates, then \textit{Wrong Prediction Flip}, insofar as where it is valid.}
\label{tab:watermark_sepcand_dup5_nonsep1_near2}
\renewcommand{\arraystretch}{0.60}
\begin{tabular}{@{\hskip 3pt}l@{\hskip 3pt}@{\hskip 3pt}c@{\hskip 3pt}c@{\hskip 3pt}c@{\hskip 3pt}c@{\hskip 3pt}c@{\hskip 3pt}c@{\hskip 3pt}c@{\hskip 3pt}c@{\hskip 3pt}c@{\hskip 3pt}c@{\hskip 3pt}c@{\hskip 3pt}c@{\hskip 3pt}c@{\hskip 3pt}c@{\hskip 3pt}c@{\hskip 3pt}c@{\hskip 3pt}c@{\hskip 3pt}c@{\hskip 3pt}|@{\hskip 3pt}c@{\hskip 3pt}c@{\hskip 3pt}c@{}}
\toprule
Dataset & \multicolumn{3}{@{\hskip -0.2pt}c}{Avila} & \multicolumn{3}{@{\hskip -0.2pt}c}{Img Seg.} & \multicolumn{3}{@{\hskip -0.2pt}c}{Letter Recognition} & \multicolumn{3}{@{\hskip -0.2pt}c}{optdigits} & \multicolumn{3}{@{\hskip -0.2pt}c}{pendigits} & \multicolumn{3}{@{\hskip -0.2pt}c}{Wine Quality} & \multicolumn{3}{@{\hskip -0.2pt}c}{average} \\
\cmidrule(l{1pt}r{6pt}){2-4} \cmidrule(l{1pt}r{6pt}){5-7} \cmidrule(l{1pt}r{6pt}){8-10} \cmidrule(l{1pt}r{6pt}){11-13} \cmidrule(l{1pt}r{6pt}){14-16} \cmidrule(l{1pt}r{6pt}){17-19} \cmidrule(l{1pt}r{6pt}){20-22}
Method | Ratio & 0.001 & 0.01 & 0.1 & 0.001 & 0.01 & 0.1 & 0.001 & 0.01 & 0.1 & 0.001 & 0.01 & 0.1 & 0.001 & 0.01 & 0.1 & 0.001 & 0.01 & 0.1 & 0.001 & 0.01 & 0.1 \\
\midrule
Cluster (Conf) & \textbf{1.000} & 0.881 & 0.871 & \textbf{1.000} & \textbf{1.000} & \textbf{1.000} & \textbf{1.000} & 0.833 & 0.930 & \textbf{1.000} & \textbf{1.000} & \textbf{1.000} & \textbf{1.000} & \textbf{1.000} & \textbf{1.000} & \textbf{1.000} & 0.850 & \textbf{1.000} & \textbf{1.000} & 0.927 & \textbf{0.967} \\
Cluster (Dist) & \textbf{1.000} & \textbf{1.000} & 0.990 & \textbf{1.000} & \textbf{1.000} & \textbf{1.000} & 0.667 & 0.517 & 0.723 & \textbf{1.000} & \textbf{1.000} & \textbf{1.000} & \textbf{1.000} & \textbf{1.000} & \textbf{1.000} & 0.500 & 0.900 & 0.933 & 0.861 & 0.903 & 0.941 \\
Outlier (Conf) & \textbf{1.000} & \textbf{1.000} & 0.967 & \textbf{1.000} & \textbf{1.000} & \textbf{1.000} & \textbf{1.000} & \textbf{1.000} & 0.538 & \textbf{1.000} & \textbf{1.000} & \textbf{1.000} & \textbf{1.000} & \textbf{1.000} & \textbf{1.000} & \textbf{1.000} & 0.950 & 0.672 & \textbf{1.000} & 0.992 & 0.863 \\
Outlier (Dist) & \textbf{1.000} & 0.976 & \textbf{0.993} & \textbf{1.000} & \textbf{1.000} & \textbf{1.000} & \textbf{1.000} & 0.633 & 0.490 & \textbf{1.000} & \textbf{1.000} & \textbf{1.000} & \textbf{1.000} & \textbf{1.000} & \textbf{1.000} & \textbf{1.000} & 0.950 & 0.682 & \textbf{1.000} & 0.927 & 0.861 \\
Wrong (Conf) & --- & --- & --- & \textbf{1.000} & \textbf{1.000} & --- & \textbf{1.000} & \textbf{1.000} & --- & \textbf{1.000} & --- & --- & \textbf{1.000} & --- & --- & \textbf{1.000} & \textbf{1.000} & --- & \textbf{1.000} & \textbf{1.000} & --- \\
Wrong (Dist) & --- & --- & --- & \textbf{1.000} & \textbf{1.000} & --- & \textbf{1.000} & \textbf{1.000} & --- & \textbf{1.000} & --- & --- & \textbf{1.000} & --- & --- & \textbf{1.000} & 0.950 & --- & \textbf{1.000} & 0.983 & --- \\
Conf. (Conf) & \textbf{1.000} & \textbf{1.000} & 0.734 & \textbf{1.000} & \textbf{1.000} & \textbf{1.000} & \textbf{1.000} & \textbf{1.000} & \textbf{0.972} & \textbf{1.000} & \textbf{1.000} & \textbf{1.000} & \textbf{1.000} & \textbf{1.000} & \textbf{1.000} & \textbf{1.000} & \textbf{1.000} & 0.974 & \textbf{1.000} & \textbf{1.000} & 0.947 \\
Conf. (Dist) & \textbf{1.000} & \textbf{1.000} & 0.828 & \textbf{1.000} & \textbf{1.000} & \textbf{1.000} & \textbf{1.000} & \textbf{1.000} & 0.880 & \textbf{1.000} & \textbf{1.000} & \textbf{1.000} & \textbf{1.000} & \textbf{1.000} & \textbf{1.000} & \textbf{1.000} & \textbf{1.000} & 0.851 & \textbf{1.000} & \textbf{1.000} & 0.927 \\
Random (Conf) & \textbf{1.000} & 0.810 & 0.670 & \textbf{1.000} & \textbf{1.000} & \textbf{1.000} & \textbf{1.000} & 0.817 & 0.487 & \textbf{1.000} & \textbf{1.000} & \textbf{1.000} & \textbf{1.000} & \textbf{1.000} & \textbf{1.000} & \textbf{1.000} & 0.850 & 0.646 & \textbf{1.000} & 0.913 & 0.800 \\
Random (Dist) & \textbf{1.000} & 0.952 & 0.888 & \textbf{1.000} & \textbf{1.000} & \textbf{1.000} & \textbf{1.000} & 0.750 & 0.495 & \textbf{1.000} & \textbf{1.000} & \textbf{1.000} & \textbf{1.000} & \textbf{1.000} & \textbf{1.000} & \textbf{1.000} & 0.800 & 0.713 & \textbf{1.000} & 0.917 & 0.849 \\
\bottomrule
\end{tabular}

\vspace{0.1em}
\centering
\scriptsize
\caption{Adjusted model accuracy for $\mathcal{D}_{\text{cand}} = \mathcal{D}_{\text{train}}$ and $\textit{d}_{cand=train} = 5$.  \textit{Cluster Flip}, \textit{Confidence Flip}, \textit{Outlier Flip} all show competitive results.}
\label{tab:adj_model_accuracy_trainonly_dup5_nonsep1_near2}
\begin{tabular}{@{\hskip 3pt}l@{\hskip 3pt}@{\hskip 3pt}c@{\hskip 3pt}c@{\hskip 3pt}c@{\hskip 3pt}c@{\hskip 3pt}c@{\hskip 3pt}c@{\hskip 3pt}c@{\hskip 3pt}c@{\hskip 3pt}c@{\hskip 3pt}c@{\hskip 3pt}c@{\hskip 3pt}c@{\hskip 3pt}c@{\hskip 3pt}c@{\hskip 3pt}c@{\hskip 3pt}c@{\hskip 3pt}c@{\hskip 3pt}c@{\hskip 3pt}|@{\hskip 3pt}c@{\hskip 3pt}c@{\hskip 3pt}c@{}}
\toprule
Dataset & \multicolumn{3}{@{\hskip -0.2pt}c}{Avila} & \multicolumn{3}{@{\hskip -0.2pt}c}{Img Seg.} & \multicolumn{3}{@{\hskip -0.2pt}c}{Letter Recognition} & \multicolumn{3}{@{\hskip -0.2pt}c}{optdigits} & \multicolumn{3}{@{\hskip -0.2pt}c}{pendigits} & \multicolumn{3}{@{\hskip -0.2pt}c}{Wine Quality} & \multicolumn{3}{@{\hskip -0.2pt}c}{average} \\
\cmidrule(l{1pt}r{6pt}){2-4} \cmidrule(l{1pt}r{6pt}){5-7} \cmidrule(l{1pt}r{6pt}){8-10} \cmidrule(l{1pt}r{6pt}){11-13} \cmidrule(l{1pt}r{6pt}){14-16} \cmidrule(l{1pt}r{6pt}){17-19} \cmidrule(l{1pt}r{6pt}){20-22}
Method | Ratio & 0.001 & 0.01 & 0.1 & 0.001 & 0.01 & 0.1 & 0.001 & 0.01 & 0.1 & 0.001 & 0.01 & 0.1 & 0.001 & 0.01 & 0.1 & 0.001 & 0.01 & 0.1 & 0.001 & 0.01 & 0.1 \\
\midrule
Cluster (Conf) & 0.998 & 0.996 & 0.939 & \textbf{0.846} & 0.846 & 0.827 & 0.967 & 0.848 & 0.928 & 0.486 & \textbf{0.977} & 0.960 & 0.243 & \textbf{0.973} & 0.961 & 0.655 & 0.641 & 0.618 & 0.699 & \textbf{0.880} & 0.872 \\
Cluster (Dist) & 0.998 & 0.998 & 0.965 & \textbf{0.846} & 0.846 & 0.827 & \textbf{0.968} & 0.489 & 0.900 & 0.487 & 0.974 & \textbf{0.965} & 0.729 & 0.970 & 0.961 & 0.655 & 0.647 & 0.609 & 0.781 & 0.820 & 0.871 \\
Outlier (Conf) & 0.998 & 0.996 & 0.950 & \textbf{0.846} & 0.846 & 0.827 & 0.849 & 0.693 & 0.917 & 0.488 & 0.971 & 0.945 & \textbf{0.973} & \textbf{0.973} & 0.940 & 0.659 & 0.645 & 0.632 & \textbf{0.802} & 0.854 & 0.869 \\
Outlier (Dist) & 0.998 & 0.996 & 0.957 & \textbf{0.846} & 0.846 & 0.827 & 0.963 & 0.732 & 0.933 & \textbf{0.974} & 0.970 & 0.947 & 0.243 & 0.972 & 0.964 & 0.664 & 0.615 & 0.637 & 0.781 & 0.855 & 0.877 \\
Wrong (Conf) & --- & --- & --- & --- & --- & --- & ---- & --- & --- & --- & --- & --- & --- & --- & --- & 0.656 & 0.639 & --- & 0.656 & 0.639 & --- \\
Wrong (Dist) & --- & --- & --- & --- & --- & --- & --- & --- & --- & --- & --- & --- & --- & --- & --- & 0.655 & 0.650 & --- & 0.655 & 0.650 & --- \\
Conf. (Conf) & 0.999 & 0.957 & 0.945 & \textbf{0.846} & 0.846 & 0.827 & 0.846 & \textbf{0.885} & 0.936 & 0.487 & 0.827 & 0.963 & 0.243 & 0.945 & 0.966 & \textbf{0.667} & \textbf{0.660} & 0.641 & 0.681 & 0.854 & 0.880 \\
Conf. (Dist) & 0.998 & 0.995 & 0.949 & \textbf{0.846} & 0.846 & 0.846 & 0.606 & 0.876 & \textbf{0.937} & 0.487 & 0.828 & 0.963 & 0.730 & 0.971 & 0.965 & \textbf{0.667} & 0.653 & 0.638 & 0.722 & 0.862 & \textbf{0.883} \\
Random (Conf) & 0.666 & 0.829 & 0.950 & \textbf{0.846} & \textbf{0.865} & \textbf{0.865} & 0.965 & 0.462 & 0.869 & 0.000 & 0.732 & 0.964 & 0.485 & 0.866 & 0.965 & 0.658 & 0.620 & \textbf{0.649} & 0.603 & 0.729 & 0.877 \\
Random (Dist) & \textbf{0.999} & \textbf{0.998} & \textbf{0.975} & \textbf{0.846} & \textbf{0.865} & 0.846 & 0.962 & 0.487 & 0.897 & 0.000 & 0.778 & 0.963 & 0.243 & 0.944 & \textbf{0.968} & 0.651 & 0.642 & 0.636 & 0.617 & 0.786 & 0.881 \\
\bottomrule
\end{tabular}

\vspace{0.1em}
\centering
\scriptsize
\caption{Adjusted model accuracy for selected watermarking rates, for $\mathcal{D}_{\text{cand}} \neq \mathcal{D}_{\text{train}}$, without $\mathcal{W}'$ duplication. \textit{Cluster Center Flip} has a relatively high accuracy.}
\label{tab:adj_model_accuracy_sepcand_dup5_nonsep1_near2}
\begin{tabular}{@{\hskip 3pt}l@{\hskip 3pt}@{\hskip 3pt}c@{\hskip 3pt}c@{\hskip 3pt}c@{\hskip 3pt}c@{\hskip 3pt}c@{\hskip 3pt}c@{\hskip 3pt}c@{\hskip 3pt}c@{\hskip 3pt}c@{\hskip 3pt}c@{\hskip 3pt}c@{\hskip 3pt}c@{\hskip 3pt}c@{\hskip 3pt}c@{\hskip 3pt}c@{\hskip 3pt}c@{\hskip 3pt}c@{\hskip 3pt}c@{\hskip 3pt}|@{\hskip 3pt}c@{\hskip 3pt}c@{\hskip 3pt}c@{}}
\toprule
Dataset & \multicolumn{3}{@{\hskip -0.2pt}c}{Avila} & \multicolumn{3}{@{\hskip -0.2pt}c}{Img Seg.} & \multicolumn{3}{@{\hskip -0.2pt}c}{Letter Recognition} & \multicolumn{3}{@{\hskip -0.2pt}c}{optdigits} & \multicolumn{3}{@{\hskip -0.2pt}c}{pendigits} & \multicolumn{3}{@{\hskip -0.2pt}c}{Wine Quality} & \multicolumn{3}{@{\hskip -0.2pt}c}{average} \\
\cmidrule(l{1pt}r{6pt}){2-4} \cmidrule(l{1pt}r{6pt}){5-7} \cmidrule(l{1pt}r{6pt}){8-10} \cmidrule(l{1pt}r{6pt}){11-13} \cmidrule(l{1pt}r{6pt}){14-16} \cmidrule(l{1pt}r{6pt}){17-19} \cmidrule(l{1pt}r{6pt}){20-22}
Method | Ratio & 0.001 & 0.01 & 0.1 & 0.001 & 0.01 & 0.1 & 0.001 & 0.01 & 0.1 & 0.001 & 0.01 & 0.1 & 0.001 & 0.01 & 0.1 & 0.001 & 0.01 & 0.1 & 0.001 & 0.01 & 0.1 \\
\midrule
Cluster (Conf) & 0.998 & 0.878 & 0.848 & \textbf{0.846} & \textbf{0.846} & \textbf{0.846} & 0.961 & 0.803 & 0.878 & 0.973 & 0.972 & 0.958 & \textbf{0.973} & 0.971 & 0.963 & 0.629 & 0.544 & \textbf{0.614} & \textbf{0.897} & 0.836 & \textbf{0.851} \\
Cluster (Dist) & 0.997 & \textbf{0.996} & 0.965 & \textbf{0.846} & \textbf{0.846} & 0.827 & 0.642 & 0.498 & 0.682 & 0.972 & 0.974 & 0.957 & 0.971 & 0.971 & 0.959 & 0.318 & 0.557 & 0.573 & 0.791 & 0.807 & 0.827 \\
Outlier (Conf) & 0.997 & 0.994 & 0.944 & 0.827 & 0.827 & 0.827 & 0.956 & 0.937 & 0.510 & 0.970 & 0.965 & 0.949 & 0.972 & 0.970 & 0.961 & 0.627 & 0.591 & 0.422 & 0.891 & 0.881 & 0.769 \\
Outlier (Dist) & 0.997 & 0.971 & \textbf{0.971} & 0.827 & 0.827 & 0.788 & 0.955 & 0.601 & 0.466 & 0.972 & 0.965 & 0.952 & 0.970 & 0.969 & 0.961 & 0.629 & 0.587 & 0.422 & 0.892 & 0.820 & 0.760 \\
Wrong (Conf) & --- & --- & --- & \textbf{0.846} & \textbf{0.846} & --- & 0.965 & 0.957 & --- & \textbf{0.975} & --- & --- & 0.971 & --- & --- & 0.632 & \textbf{0.637} & --- & 0.878 & 0.813 & --- \\
Wrong (Dist) & --- & --- & --- & \textbf{0.846} & \textbf{0.846} & --- & \textbf{0.966} & \textbf{0.961} & --- & \textbf{0.975} & --- & --- & 0.969 & --- & --- & \textbf{0.639} & 0.597 & --- & 0.879 & 0.801 & --- \\
Conf. (Conf) & 0.997 & 0.992 & 0.708 & 0.827 & 0.827 & 0.827 & 0.963 & 0.957 & \textbf{0.915} & 0.971 & 0.972 & 0.967 & 0.971 & 0.969 & 0.965 & 0.631 & 0.623 & 0.598 & 0.893 & 0.890 & 0.830 \\
Conf. (Dist) & \textbf{0.998} & 0.995 & 0.807 & 0.827 & 0.827 & 0.827 & 0.964 & 0.956 & 0.831 & 0.969 & 0.969 & 0.968 & 0.971 & 0.970 & 0.961 & 0.631 & 0.634 & 0.520 & 0.893 & \textbf{0.892} & 0.819 \\
Random (Conf) & 0.997 & 0.807 & 0.660 & 0.827 & 0.827 & 0.827 & 0.958 & 0.778 & 0.465 & 0.974 & \textbf{0.975} & 0.965 & 0.969 & 0.973 & 0.959 & 0.633 & 0.533 & 0.408 & 0.893 & 0.815 & 0.714 \\
Random (Dist) & 0.996 & 0.949 & 0.873 & 0.827 & 0.827 & 0.827 & 0.956 & 0.710 & 0.471 & 0.974 & 0.973 & \textbf{0.969} & 0.970 & \textbf{0.973} & \textbf{0.965} & 0.637 & 0.502 & 0.438 & 0.893 & 0.822 & 0.757 \\
\bottomrule
\end{tabular}

\vspace{0.1em}
\centering
\scriptsize
\caption{Robustness for $\mathcal{D}_{\text{cand}} = \mathcal{D}_{\text{train}}$ and $\textit{d}_{cand=train} = 5$. All proposed methods show good performance, although the results at $|\mathcal{W}|/|\mathcal{D}_{train}| = 0.01$ show higher variance, likely due to the small number of watermarks involved. \textit{Confidence Flip} is overall slightly more resilent than other methods, although the difference is limited.}
\label{tab:adj_robustness_trainonly_dup5_nonsep1_near2}
\begin{tabular}{@{\hskip 3pt}l@{\hskip 3pt}@{\hskip 3pt}c@{\hskip 3pt}c@{\hskip 3pt}c@{\hskip 3pt}c@{\hskip 3pt}c@{\hskip 3pt}c@{\hskip 3pt}c@{\hskip 3pt}c@{\hskip 3pt}c@{\hskip 3pt}c@{\hskip 3pt}c@{\hskip 3pt}c@{\hskip 3pt}c@{\hskip 3pt}c@{\hskip 3pt}c@{\hskip 3pt}c@{\hskip 3pt}c@{\hskip 3pt}c@{\hskip 3pt}|@{\hskip 3pt}c@{\hskip 3pt}c@{\hskip 3pt}c@{}}
\toprule
Dataset & \multicolumn{3}{@{\hskip -0.2pt}c}{Avila} & \multicolumn{3}{@{\hskip -0.2pt}c}{Img Seg.} & \multicolumn{3}{@{\hskip -0.2pt}c}{Letter Recognition} & \multicolumn{3}{@{\hskip -0.2pt}c}{optdigits} & \multicolumn{3}{@{\hskip -0.2pt}c}{pendigits} & \multicolumn{3}{@{\hskip -0.2pt}c}{Wine Quality} & \multicolumn{3}{@{\hskip -0.2pt}c}{average} \\
\cmidrule(l{1pt}r{6pt}){2-4} \cmidrule(l{1pt}r{6pt}){5-7} \cmidrule(l{1pt}r{6pt}){8-10} \cmidrule(l{1pt}r{6pt}){11-13} \cmidrule(l{1pt}r{6pt}){14-16} \cmidrule(l{1pt}r{6pt}){17-19} \cmidrule(l{1pt}r{6pt}){20-22}
Method | Ratio & 0.001 & 0.01 & 0.1 & 0.001 & 0.01 & 0.1 & 0.001 & 0.01 & 0.1 & 0.001 & 0.01 & 0.1 & 0.001 & 0.01 & 0.1 & 0.001 & 0.01 & 0.1 & 0.001 & 0.01 & 0.1 \\
\midrule
Cluster (Conf) & \textbf{1.000} & \textbf{1.000} & 0.989 & \textbf{1.000} & \textbf{1.000} & \textbf{1.000} & 0.750 & 0.747 & 0.781 & 0.500 & \textbf{1.000} & \textbf{1.000} & \textbf{1.000} & \textbf{1.000} & \textbf{1.000} & \textbf{1.000} & \textbf{1.000} & \textbf{1.000} & 0.875 & 0.958 & 0.962 \\
Cluster (Dist) & \textbf{1.000} & \textbf{1.000} & \textbf{1.000} & \textbf{1.000} & \textbf{1.000} & \textbf{1.000} & 0.875 & 0.400 & 0.585 & \textbf{1.000} & \textbf{1.000} & \textbf{1.000} & \textbf{1.000} & \textbf{1.000} & \textbf{1.000} & \textbf{1.000} & \textbf{1.000} & 0.980 & 0.979 & 0.900 & 0.927 \\
Outlier (Conf) & \textbf{1.000} & \textbf{1.000} & \textbf{1.000} & \textbf{1.000} & \textbf{1.000} & \textbf{1.000} & 0.625 & 0.560 & 0.755 & 0.000 & 0.950 & \textbf{1.000} & 0.250 & \textbf{1.000} & \textbf{1.000} & \textbf{1.000} & \textbf{1.000} & \textbf{1.000} & 0.646 & 0.918 & 0.959 \\
Outlier (Dist) & \textbf{1.000} & \textbf{1.000} & \textbf{1.000} & \textbf{1.000} & \textbf{1.000} & \textbf{1.000} & \textbf{1.000} & 0.547 & 0.540 & 0.000 & 0.800 & \textbf{1.000} & \textbf{1.000} & \textbf{1.000} & \textbf{1.000} & \textbf{1.000} & \textbf{1.000} & \textbf{1.000} & 0.833 & 0.891 & 0.923 \\
Wrong (Conf) & --- & --- & --- & --- & --- & --- & --- & --- & --- & --- & --- & --- & --- & --- & --- & \textbf{1.000} & \textbf{1.000} & --- & \textbf{1.000} & \textbf{1.000} & - \\
Wrong (Dist) & --- & --- & --- & --- & --- & --- & --- & --- & --- & --- & --- & --- & --- & --- & --- & \textbf{1.000} & \textbf{1.000} & --- & \textbf{1.000} & \textbf{1.000} & - \\
Conf. (Conf) & \textbf{1.000} & \textbf{1.000} & 0.998 & \textbf{1.000} & \textbf{1.000} & \textbf{1.000} & \textbf{1.000} & \textbf{0.960} & 0.919 & 0.500 & 0.900 & \textbf{1.000} & 0.500 & 0.947 & \textbf{1.000} & \textbf{1.000} & \textbf{1.000} & \textbf{1.000} & 0.833 & 0.968 & 0.986 \\
Conf. (Dist) & \textbf{1.000} & \textbf{1.000} & 0.990 & \textbf{1.000} & \textbf{1.000} & \textbf{1.000} & \textbf{1.000} & 0.907 & \textbf{0.947} & 0.500 & 0.950 & 0.995 & 0.500 & \textbf{1.000} & \textbf{1.000} & \textbf{1.000} & \textbf{1.000} & \textbf{1.000} & 0.833 & 0.976 & \textbf{0.989} \\
Random (Conf) & 0.667 & 0.962 & 0.939 & \textbf{1.000} & \textbf{1.000} & \textbf{1.000} & 0.750 & 0.467 & 0.605 & 0.500 & 0.800 & \textbf{1.000} & 0.750 & \textbf{1.000} & \textbf{1.000} & \textbf{1.000} & 0.960 & 0.992 & 0.778 & 0.865 & 0.923 \\
Random (Dist) & \textbf{1.000} & 0.981 & 0.989 & \textbf{1.000} & \textbf{1.000} & \textbf{1.000} & 0.750 & 0.400 & 0.616 & 0.000 & 0.850 & \textbf{1.000} & \textbf{1.000} & \textbf{1.000} & \textbf{1.000} & \textbf{1.000} & \textbf{1.000} & \textbf{1.000} & 0.792 & 0.872 & 0.934 \\
\bottomrule
\end{tabular}

\vspace{0.1em}
\centering
\scriptsize
\caption{Robustness for $\mathcal{D}_{\text{cand}} \neq \mathcal{D}_{\text{train}}$. Similarly to Table~\ref{tab:adj_robustness_trainonly_dup5_nonsep1_near2},  the results at $|\mathcal{W}|/|\mathcal{D}_{train}| = 0.01$ show high variance.}
\label{tab:adj_robustness_sepcand_dup5_nonsep1_near2}
\begin{tabular}{@{\hskip 3pt}l@{\hskip 3pt}@{\hskip 3pt}c@{\hskip 3pt}c@{\hskip 3pt}c@{\hskip 3pt}c@{\hskip 3pt}c@{\hskip 3pt}c@{\hskip 3pt}c@{\hskip 3pt}c@{\hskip 3pt}c@{\hskip 3pt}c@{\hskip 3pt}c@{\hskip 3pt}c@{\hskip 3pt}c@{\hskip 3pt}c@{\hskip 3pt}c@{\hskip 3pt}c@{\hskip 3pt}c@{\hskip 3pt}c@{\hskip 3pt}|@{\hskip 3pt}c@{\hskip 3pt}c@{\hskip 3pt}c@{}}
\toprule
Dataset & \multicolumn{3}{@{\hskip -0.2pt}c}{Avila} & \multicolumn{3}{@{\hskip -0.2pt}c}{Img Seg.} & \multicolumn{3}{@{\hskip -0.2pt}c}{Letter Recognition} & \multicolumn{3}{@{\hskip -0.2pt}c}{optdigits} & \multicolumn{3}{@{\hskip -0.2pt}c}{pendigits} & \multicolumn{3}{@{\hskip -0.2pt}c}{Wine Quality} & \multicolumn{3}{@{\hskip -0.2pt}c}{average} \\
\cmidrule(l{1pt}r{6pt}){2-4} \cmidrule(l{1pt}r{6pt}){5-7} \cmidrule(l{1pt}r{6pt}){8-10} \cmidrule(l{1pt}r{6pt}){11-13} \cmidrule(l{1pt}r{6pt}){14-16} \cmidrule(l{1pt}r{6pt}){17-19} \cmidrule(l{1pt}r{6pt}){20-22}
Method | Ratio & 0.001 & 0.01 & 0.1 & 0.001 & 0.01 & 0.1 & 0.001 & 0.01 & 0.1 & 0.001 & 0.01 & 0.1 & 0.001 & 0.01 & 0.1 & 0.001 & 0.01 & 0.1 & 0.001 & 0.01 & 0.1 \\
\midrule
Cluster (Conf) & \textbf{1.000} & 0.860 & 0.717 & \textbf{1.000} & \textbf{1.000} & \textbf{1.000} & \textbf{1.000} & 0.556 & 0.811 & \textbf{1.000} & \textbf{1.000} & \textbf{1.000} & \textbf{1.000} & \textbf{1.000} & \textbf{1.000} & \textbf{1.000} & 0.765 & \textbf{0.979} & \textbf{1.000} & 0.863 & \textbf{0.918} \\
Cluster (Dist) & \textbf{1.000} & \textbf{1.000} & \textbf{0.983} & \textbf{1.000} & \textbf{1.000} & \textbf{1.000} & 0.444 & 0.233 & 0.450 & \textbf{1.000} & \textbf{1.000} & \textbf{1.000} & \textbf{1.000} & \textbf{1.000} & \textbf{1.000} & 0.250 & 0.855 & 0.871 & 0.782 & 0.848 & 0.884 \\
Outlier (Conf) & \textbf{1.000} & 0.952 & 0.913 & \textbf{1.000} & \textbf{1.000} & \textbf{1.000} & 0.333 & 0.433 & 0.240 & \textbf{1.000} & \textbf{1.000} & \textbf{1.000} & \textbf{1.000} & \textbf{1.000} & \textbf{1.000} & \textbf{1.000} & 0.760 & 0.462 & 0.889 & 0.858 & 0.769 \\
Outlier (Dist) & \textbf{1.000} & 0.976 & 0.967 & \textbf{1.000} & \textbf{1.000} & \textbf{1.000} & 0.667 & 0.348 & 0.194 & \textbf{1.000} & \textbf{1.000} & \textbf{1.000} & \textbf{1.000} & \textbf{1.000} & \textbf{1.000} & \textbf{1.000} & 0.760 & 0.462 & 0.944 & 0.847 & 0.770 \\
Wrong (Conf) & --- & --- & --- & \textbf{1.000} & \textbf{1.000} & --- & \textbf{1.000} & 0.967 & --- & \textbf{1.000} & --- & --- & \textbf{1.000} & --- & --- & \textbf{1.000} & 0.950 & --- & \textbf{1.000} & 0.972 & --- \\
Wrong (Dist) & --- & --- & --- & \textbf{1.000} & \textbf{1.000} & --- & \textbf{1.000} & 0.950 & --- & \textbf{1.000} & --- & --- & \textbf{1.000} & --- & --- & \textbf{1.000} & 0.902 & --- & \textbf{1.000} & 0.951 & --- \\
Conf. (Conf) & \textbf{1.000} & 0.833 & 0.525 & \textbf{1.000} & \textbf{1.000} & \textbf{1.000} & \textbf{1.000} & \textbf{1.000} & \textbf{0.881} & \textbf{1.000} & \textbf{1.000} & \textbf{1.000} & \textbf{1.000} & \textbf{1.000} & \textbf{1.000} & \textbf{1.000} & \textbf{1.000} & 0.919 & \textbf{1.000} & 0.972 & 0.888 \\
Conf. (Dist) & \textbf{1.000} & \textbf{1.000} & 0.620 & \textbf{1.000} & \textbf{1.000} & \textbf{1.000} & \textbf{1.000} & \textbf{1.000} & 0.697 & \textbf{1.000} & \textbf{1.000} & \textbf{1.000} & \textbf{1.000} & \textbf{1.000} & \textbf{1.000} & \textbf{1.000} & \textbf{1.000} & 0.712 & \textbf{1.000} & \textbf{1.000} & 0.838 \\
Random (Conf) & 0.800 & 0.636 & 0.442 & \textbf{1.000} & \textbf{1.000} & \textbf{1.000} & 0.333 & 0.381 & 0.212 & \textbf{1.000} & \textbf{1.000} & \textbf{1.000} & \textbf{1.000} & \textbf{1.000} & \textbf{1.000} & 0.500 & 0.680 & 0.447 & 0.772 & 0.783 & 0.684 \\
Random (Dist) & \textbf{1.000} & 0.952 & 0.747 & \textbf{1.000} & \textbf{1.000} & \textbf{1.000} & 0.667 & 0.375 & 0.214 & \textbf{1.000} & \textbf{1.000} & \textbf{1.000} & \textbf{1.000} & \textbf{1.000} & \textbf{1.000} & \textbf{1.000} & 0.560 & 0.483 & 0.944 & 0.815 & 0.741 \\
\bottomrule
\end{tabular}

\end{table*}

\begin{table*}[t]
    \centering
    \small
    \caption{Summary of data contexts, empirical performance, and proposed mechanisms based on experimental results in Section~\ref{sec:exp}.}
    \label{tab:selection-guidance}
    \begin{tabular}{|p{3cm}|p{3.5cm}|p{3.5cm}|p{5.3cm}|}
        \hline
        \textbf{Strategy} & \textbf{Recommended Context} & \textbf{Performance} & \textbf{Proposed Mechanism} \\
        \hline
        \textbf{Cluster Center Flip} & $\mathcal{D}_{cand} \neq \mathcal{D}_{train}$ (i.e. the watermarking dataset is separate from the training dataset) & Shows high adjusted model accuracy (Table~\ref{tab:adj_model_accuracy_sepcand_dup5_nonsep1_near2}) and good effectiveness and confidence (Table~\ref{tab:watermark_sepcand_dup5_nonsep1_near2},~\ref{tab:adj_robustness_sepcand_dup5_nonsep1_near2}) w.r.t. $\mathcal{D}_{cand} \neq \mathcal{D}_{train}$ and Lowest Confidence candidate selection (see Section~\ref{sec:candidate_selection}) & $\mathcal{D}_{\text{cand}} \neq \mathcal{D}_{\text{train}}$ (and Lowest Confidence selection as in Section~\ref{sec:candidate_selection}) helps watermarks avoid entanglement between the candidate clusters and confident learned boundaries, which helps anchoring minimize the effect of the embedding on non-watermark samples, as well as make distancing clusters less important.\\
        \hline
        \textbf{Confidence Flip} & $\mathcal{D}_{cand} = \mathcal{D}_{train}$ (i.e. the watermarking dataset overlaps with the training dataset). & Shows good relative performance, especially for $\mathcal{D}_{cand} = \mathcal{D}_{train}$. & Targeting lower confidence samples eases the effort in embedding the sample, and thus leads to better effectiveness and resilience.\\ 

        \hline
        \textbf{Outlier Flip} & Low Watermarking Rates (e.g., $|\mathcal{W}|/|\mathcal{D}_{train}| = 0.001$). & Generally effective at low rates, but less competitive at higher rates (e.g. $|\mathcal{W}|/|\mathcal{D}_{train}| = 0.1$). & Spatial isolation of "outlier" samples is lost at high watermarking rates, reducing effectiveness. \\
        \hline
        \textbf{Wrong Prediction Flip} & Datasets with enough wrongly predicted samples for a valid $\mathcal{D}_{cand}$. & Good performance w.r.t. most metrics when viable. & Only viable when a large supply of misclassified samples is abundant, which is heavily dataset-dependent - and generally only $\mathcal{D}_{cand} \neq \mathcal{D}_{train}$ (see validity between Table~\ref{tab:watermark_sepcand_dup5_nonsep1_near2} and Table~\ref{tab:watermark_trainonly_dup5_nonsep1_near2}) due to tendency of GBDT to overfit to training dataset.\\
        \hline
    \end{tabular}
\end{table*}

\bibliography{aaai2026}

\newpage

\appendix
\section{Appendix}

\subsection{Dataset details} \label{sec:dataset}

The datasets used in Section 4 are detailed in Table~\ref{tab:dataset_stats}.

\begin{table*}[h]
\centering
\caption{Dataset statistics as used in Section 4.}
\label{tab:dataset_stats}
\begin{tabular}{lrrrr}
\toprule
\midrule
Dataset            & Features & Classes & Train Size & Test Size  \\
\midrule
Avila~\cite{avila459} & 10 & 12 & 10430 & 10437 \\
Image Segmentation (Img Seg.)~\cite{imagesegmentation50} & 19       & 7       & 157       & 53        \\
Letter Recognition~\cite{letterrecognition59} & 16       & 26      & 15000      & 5000       \\
Wine Quality~\cite{DBLP:journals/dss/CortezCAMR09}            & 13       & 6       & 4872        & 1625        \\
optdigits~\cite{DBLP:journals/tsmc/XuKS92}          & 64       & 10      & 3812       & 1796       \\
pendigits~\cite{pendigits81}          & 16       & 10      & 7483       & 3497       \\
\midrule
\bottomrule
\end{tabular}
\end{table*}

\subsection{Watermark Selection} \label{sec:watermark_selection_appendix}
\begin{algorithm}[htpb]
 \caption{Watermark Candidate Selection via Maximum Distance}
\label{alg:max_dist}
\algsetup{linenodelimiter=.}
\raggedright \textbf{Input}: Set of watermark candidates $\mathcal{C} = \{c_1, c_2, \dots, c_n\}$ where $c_i = (\mathbf{x}_i, y_i)$, desired number of watermarks $k$\\
\raggedright \textbf{Output}: Selected watermark set $\mathcal{W} \subset \mathcal{C}$, $|\mathcal{W}| = k$
\begin{algorithmic}[1]

\STATE Initialize $\mathcal{W} = \{\}$

\STATE Select an initial sample $c^*$ from $\mathcal{C}$ randomly
\STATE $\mathcal{W} \gets \mathcal{W} \cup \{c^*\}$

\FOR{$i = 2$ to $k$}
    \FOR{each $c \in \mathcal{C} \setminus \mathcal{W}$}
        \STATE Compute $\text{min\_dist}(c) = \min_{w \in \mathcal{W}} \text{dist}(c, w)$
    \ENDFOR
    \STATE Select $c' = \arg\max_{c \in \mathcal{C}\setminus \mathcal{W}} \text{min\_dist}(c)$
    \STATE $\mathcal{W} \gets \mathcal{W} \cup \{c'\}$
\ENDFOR

\end{algorithmic}
\end{algorithm}

\subsection{Candidate Resilience} \label{sec:cand_resil}
In order to encode information into the watermark, It is necessary to define a set group of candidate samples which can be modified or kept as-is to signify a value of 0 or 1. It is also important for those samples to be independent as possible, in order to prevent non-modified samples being flipped from embedding due to influence from the active watermark samples, which may obfuscate the information embedded in the watermark. We test this independence (or resilience of candidates to watermarking w.r.t. other samples) as follows: as seen in the process described in Section 4.1 we first obtain an initial candidate set of size \textit{n}, from which fine-tune the initial model (that was training using the $\mathcal{D}_{\text{train}}$ dataset) to apply a watermark to \textit{k} samples. 

We quantify the candidate resilience $\mathcal{R}_{\text{cand}}$ by the proportion of remaining \textit{n-k} samples that the watermarked model has the same predicted label as the initial model:
\setlength{\abovedisplayskip}{1pt}
\setlength{\belowdisplayskip}{1pt}
\begin{align}
\mathcal{R}_{\text{cand}} = \frac{1}{|\mathcal{C} \setminus \mathcal{W}|} \sum_{(\mathbf{x}_i, y_i) \in \mathcal{C} \setminus \mathcal{W}} \mathbf{1} (F^{\text{wm}}(\mathbf{x}_i) = y_i)
\label{eq:cand_res}
\end{align}
Similarly to general accuracy, we report the adjusted candidate resilience, defined as $\mathcal{R}_{\text{cand}}' = \mathcal{A}_{\text{wm}} \cdot \mathcal{R}_{\text{cand}}$, to account for the fact that resilience is only meaningful when the watermark is successfully embedded.

Table~\ref{tab:adj_resilience_trainonly_dup5_nonsep1_near2} shows results for \( \mathcal{D}_{\text{cand}} = \mathcal{D}_{\text{train}} \), while Table~\ref{tab:adj_resilience_sepcand_dup5_nonsep1_near2} presents the separate-candidate setting. \textit{Cluster Center Flip}, which selects centroids from distinct clusters, thus implicitly favoring more distant points, achieves the highest resilience scores, though other methods also perform reasonably well.

% Candidate sample resilience results are shown in Figure~\ref{fig:candidate_resilience}. As is apparent, for watermark candidate numbers that within reasonable range, the flip rate is generally zero, which is necessary for any form of reliable watermark.
\begin{table*}[thbp]
\centering
\setlength{\abovecaptionskip}{1pt}
\setlength{\belowcaptionskip}{1pt}
\tiny
\caption{Adjusted resilience for $\mathcal{D}_{\text{cand}} = \mathcal{D}_{\text{train}}$ and $\textit{d}_{cand=train} = 5$. Resilience is generally high for all methods, especially for \textit{Cluster Center Flip} (with the exception of \textit{Wrong Prediction Flip}, which is only valid for one dataset). \textit{Random Flip} baseline being lower than the other methods by a significant gap. Note that for the Image Segmentation dataset, $|\mathcal{W}|/|\mathcal{D}_{train}| = 0.01$ is invalid due to $n = 1 \text{ and } k = 1$ i.e. the size of the dataset is small enough that the integer ceiling for both is $1$.}

\label{tab:adj_resilience_trainonly_dup5_nonsep1_near2}
\begin{tabular}{@{\hskip 3pt}l@{\hskip 3pt}@{\hskip 3pt}c@{\hskip 3pt}c@{\hskip 3pt}c@{\hskip 3pt}c@{\hskip 3pt}c@{\hskip 3pt}c@{\hskip 3pt}c@{\hskip 3pt}c@{\hskip 3pt}c@{\hskip 3pt}c@{\hskip 3pt}c@{\hskip 3pt}c@{\hskip 3pt}c@{\hskip 3pt}c@{\hskip 3pt}c@{\hskip 3pt}c@{\hskip 3pt}c@{\hskip 3pt}c@{\hskip 3pt}|@{\hskip 3pt}c@{\hskip 3pt}c@{\hskip 3pt}c@{}}
\toprule
Dataset & \multicolumn{3}{@{\hskip -0.2pt}c}{Avila} & \multicolumn{3}{@{\hskip -0.2pt}c}{Img Seg.} & \multicolumn{3}{@{\hskip -0.2pt}c}{Letter Recognition} & \multicolumn{3}{@{\hskip -0.2pt}c}{optdigits} & \multicolumn{3}{@{\hskip -0.2pt}c}{pendigits} & \multicolumn{3}{@{\hskip -0.2pt}c}{Wine Quality} & \multicolumn{3}{@{\hskip -0.2pt}c}{average} \\
\cmidrule(l{1pt}r{6pt}){2-4} \cmidrule(l{1pt}r{6pt}){5-7} \cmidrule(l{1pt}r{6pt}){8-10} \cmidrule(l{1pt}r{6pt}){11-13} \cmidrule(l{1pt}r{6pt}){14-16} \cmidrule(l{1pt}r{6pt}){17-19} \cmidrule(l{1pt}r{6pt}){20-22}
Method | Ratio & 0.001 & 0.01 & 0.1 & 0.001 & 0.01 & 0.1 & 0.001 & 0.01 & 0.1 & 0.001 & 0.01 & 0.1 & 0.001 & 0.01 & 0.1 & 0.001 & 0.01 & 0.1 & 0.001 & 0.01 & 0.1 \\
\midrule
Cluster (Conf) & \textbf{1.000} & \textbf{1.000} & 0.996 & --- & \textbf{1.000} & \textbf{1.000} & \textbf{1.000} & 0.880 & 0.993 & 0.500 & \textbf{1.000} & \textbf{1.000} & 0.250 & \textbf{1.000} & \textbf{1.000} & \textbf{1.000} & \textbf{1.000} & \textbf{1.000} & 0.750 & \textbf{0.980} & \textbf{0.998} \\
Cluster (Dist) & \textbf{1.000} & \textbf{1.000} & \textbf{1.000} & --- & \textbf{1.000} & \textbf{1.000} & \textbf{1.000} & 0.507 & 0.968 & 0.500 & \textbf{1.000} & \textbf{1.000} & 0.750 & \textbf{1.000} & \textbf{1.000} & \textbf{1.000} & 0.917 & 0.918 & 0.850 & 0.904 & 0.981 \\
Outlier (Conf) & 0.800 & \textbf{1.000} & 0.998 & --- & \textbf{1.000} & \textbf{1.000} & 0.750 & 0.509 & 0.983 & 0.500 & \textbf{1.000} & \textbf{1.000} & \textbf{1.000} & \textbf{1.000} & \textbf{1.000} & \textbf{1.000} & \textbf{1.000} & 0.955 & 0.810 & 0.918 & 0.989 \\
Outlier (Dist) & 0.600 & \textbf{1.000} & 0.994 & --- & \textbf{1.000} & \textbf{1.000} & 0.143 & 0.497 & 0.993 & \textbf{1.000} & \textbf{1.000} & \textbf{1.000} & 0.250 & \textbf{1.000} & \textbf{1.000} & \textbf{1.000} & 0.880 & 0.963 & 0.599 & 0.896 & 0.992 \\
Wrong (Conf) & --- & --- & --- & --- & --- & --- & --- & --- & --- & --- & --- & --- & --- & --- & --- & 0.500 &  0.291 & --- & 0.500 &  0.291 & --- \\
Wrong (Dist) & --- & --- & --- & --- & --- & --- & --- & --- & --- & --- & --- & --- & --- & --- & --- & \textbf{1.000} & 0.417 & --- & \textbf{1.000} & 0.417 & --- \\
Conf. (Conf) & \textbf{1.000} & 0.962 & 0.975 & --- & \textbf{1.000} & \textbf{1.000} & 0.875 & \textbf{0.920} & 0.985 & 0.500 & 0.850 & \textbf{1.000} & 0.250 & 0.974 & \textbf{1.000} & 0.500 & 0.583 & 0.873 & 0.625 & 0.882 & 0.972 \\
Conf. (Dist) & \textbf{1.000} & \textbf{1.000} & 0.946 & --- & \textbf{1.000} & \textbf{1.000} & 0.625 & 0.907 & \textbf{0.993} & 0.500 & 0.850 & \textbf{1.000} & 0.750 & \textbf{1.000} & \textbf{1.000} & 0.500 & 0.667 & 0.848 & 0.675 & 0.904 & 0.965 \\
Random (Conf) & 0.667 & 0.830 & 0.971 & --- & \textbf{1.000} & \textbf{1.000} & \textbf{1.000} & 0.480 & 0.911 & 0.000 & 0.750 & \textbf{1.000} & 0.500 & 0.895 & \textbf{1.000} & \textbf{1.000} & 0.960 & 0.951 & 0.633 & 0.819 & 0.972 \\
Random (Dist) & \textbf{1.000} & \textbf{1.000} & 0.996 & --- & \textbf{1.000} & \textbf{1.000} & \textbf{1.000} & 0.507 & 0.952 & 0.000 & 0.800 & \textbf{1.000} & 0.250 & 0.974 & \textbf{1.000} & \textbf{1.000} & \textbf{1.000} & 0.988 & 0.650 & 0.880 & 0.989 \\
\bottomrule
\end{tabular}
\end{table*}

\begin{table*}[thbp]
\centering
\setlength{\abovecaptionskip}{1pt}
\setlength{\belowcaptionskip}{1pt}
\tiny
\caption{Adjusted resilience for $\mathcal{D}_{\text{cand}} \neq \mathcal{D}_{\text{train}}$, without $\mathcal{W}'$ duplication. \textit{Cluster Center Flip} shows generally superior results, although less so than for Table~\ref{tab:adj_resilience_trainonly_dup5_nonsep1_near2}.}
\label{tab:adj_resilience_sepcand_dup5_nonsep1_near2}
\begin{tabular}{@{\hskip 3pt}l@{\hskip 3pt}@{\hskip 3pt}c@{\hskip 3pt}c@{\hskip 3pt}c@{\hskip 3pt}c@{\hskip 3pt}c@{\hskip 3pt}c@{\hskip 3pt}c@{\hskip 3pt}c@{\hskip 3pt}c@{\hskip 3pt}c@{\hskip 3pt}c@{\hskip 3pt}c@{\hskip 3pt}c@{\hskip 3pt}c@{\hskip 3pt}c@{\hskip 3pt}c@{\hskip 3pt}c@{\hskip 3pt}c@{\hskip 3pt}|@{\hskip 3pt}c@{\hskip 3pt}c@{\hskip 3pt}c@{}}
\toprule
Dataset & \multicolumn{3}{@{\hskip -0.2pt}c}{Avila} & \multicolumn{3}{@{\hskip -0.2pt}c}{Img Seg.} & \multicolumn{3}{@{\hskip -0.2pt}c}{Letter Recognition} & \multicolumn{3}{@{\hskip -0.2pt}c}{optdigits} & \multicolumn{3}{@{\hskip -0.2pt}c}{pendigits} & \multicolumn{3}{@{\hskip -0.2pt}c}{Wine Quality} & \multicolumn{3}{@{\hskip -0.2pt}c}{average} \\
\cmidrule(l{1pt}r{6pt}){2-4} \cmidrule(l{1pt}r{6pt}){5-7} \cmidrule(l{1pt}r{6pt}){8-10} \cmidrule(l{1pt}r{6pt}){11-13} \cmidrule(l{1pt}r{6pt}){14-16} \cmidrule(l{1pt}r{6pt}){17-19} \cmidrule(l{1pt}r{6pt}){20-22}
Method | Ratio & 0.001 & 0.01 & 0.1 & 0.001 & 0.01 & 0.1 & 0.001 & 0.01 & 0.1 & 0.001 & 0.01 & 0.1 & 0.001 & 0.01 & 0.1 & 0.001 & 0.01 & 0.1 & 0.001 & 0.01 & 0.1 \\
\midrule
Cluster (Conf) & \textbf{1.000} & 0.881 & 0.867 & --- & \textbf{1.000} & \textbf{1.000} & \textbf{1.000} & \textbf{0.833} & 0.930 & \textbf{1.000} & \textbf{1.000} & \textbf{1.000} & \textbf{1.000} & \textbf{1.000} & 0.993 & \textbf{1.000} & \textbf{0.850} & \textbf{0.995} & \textbf{1.000} & 0.927 & \textbf{0.964} \\
Cluster (Dist) & \textbf{1.000} & \textbf{1.000} & 0.986 & --- & \textbf{1.000} & \textbf{1.000} & 0.667 & 0.499 & 0.697 & \textbf{1.000} & \textbf{1.000} & 0.987 & \textbf{1.000} & \textbf{1.000} & 0.997 & 0.500 & 0.711 & 0.866 & 0.833 & 0.868 & 0.922 \\
Outlier (Conf) & \textbf{1.000} & \textbf{1.000} & 0.957 & --- & \textbf{1.000} & 0.833 & 0.500 & 0.783 & 0.531 & \textbf{1.000} & \textbf{1.000} & \textbf{1.000} & \textbf{1.000} & \textbf{1.000} & 0.993 & 0.500 & \textbf{0.850} & 0.624 & 0.800 & \textbf{0.939} & 0.823 \\
Outlier (Dist) & \textbf{1.000} & 0.953 & \textbf{0.990} & --- & \textbf{1.000} & \textbf{1.000} & 0.000 & 0.559 & 0.483 & \textbf{1.000} & \textbf{1.000} & 0.987 & \textbf{1.000} & \textbf{1.000} & 0.970 & 0.500 & \textbf{0.850} & 0.626 & 0.700 & 0.894 & 0.843 \\
Wrong (Conf) & --- & --- & --- & --- & \textbf{1.000} & --- & 0.667 & 0.833 & --- & \textbf{1.000} & --- & --- & \textbf{1.000} & --- & --- & 0.500 & 0.579 & --- & 0.791 & 0.804 & --- \\
Wrong (Dist) & --- & --- & --- & --- & \textbf{1.000} & --- & 0.000 & 0.767 & --- & \textbf{1.000} & --- & --- & \textbf{1.000} & --- & --- & 0.000 & 0.474 & --- & 0.500 & 0.621 & --- \\
Conf. (Conf) & \textbf{1.000} & 0.929 & 0.726 & --- & \textbf{1.000} & \textbf{1.000} & 0.500 & \textbf{0.833} & \textbf{0.964} & \textbf{1.000} & \textbf{1.000} & \textbf{1.000} & \textbf{1.000} & \textbf{1.000} & 0.990 & 0.500 & 0.684 & 0.949 & 0.800 & 0.908 & 0.938 \\
Conf. (Dist) & \textbf{1.000} & \textbf{1.000} & 0.824 & --- & \textbf{1.000} & \textbf{1.000} & 0.667 & 0.633 & 0.839 & \textbf{1.000} & 0.933 & 0.993 & \textbf{1.000} & 0.967 & 0.993 & 0.500 & 0.579 & 0.747 & 0.833 & 0.852 & 0.899 \\
Random (Conf) & \textbf{1.000} & 0.810 & 0.660 & --- & \textbf{1.000} & \textbf{1.000} & \textbf{1.000} & 0.789 & 0.474 & \textbf{1.000} & \textbf{1.000} & \textbf{1.000} & \textbf{1.000} & \textbf{1.000} & \textbf{1.000} & 0.500 & 0.805 & 0.563 & 0.900 & 0.901 & 0.783 \\
Random (Dist) & \textbf{1.000} & 0.952 & 0.888 & --- & \textbf{1.000} & \textbf{1.000} & \textbf{1.000} & 0.737 & 0.483 & \textbf{1.000} & \textbf{1.000} & 0.987 & \textbf{1.000} & \textbf{1.000} & 0.997 & \textbf{1.000} & 0.800 & 0.658 & \textbf{1.000} & 0.915 & 0.835 \\
\bottomrule
\end{tabular}
\end{table*}

\subsection{Hyperparameters} \label{sec:rep_check}
We reproduce the hyperparameters for the experiments in Section 4 in Tables~\ref{tab:gbdt_hyperparameters} and~\ref{tab:watermarking_hyperparameters}. Section 4 provides additional description of those variables.

\begin{table*}[t]
\small
    \centering
    \caption{Standard GBDT model hyperparameters as used in experiments detailed in Section 4.}
    \label{tab:gbdt_hyperparameters}
    \begin{tabular}{lc}
    \toprule
    \textbf{Parameter} & \textbf{Value}  \\
    \midrule
    Iterations & 200\\
    Shrinkage & 0.1 \\
    Max Leaves & 20  \\
    Feature Sampling & 0.1\\
    \bottomrule
    \end{tabular}
\end{table*}

\begin{table*}[t]
    \centering
\small
    \caption{Watermarking/data splitting parameters as used in the  experiments detailed in Section 4.}
    \label{tab:watermarking_hyperparameters}
    \begin{tabular}{lcl}
    \toprule
    \textbf{Parameter} & \textbf{Value} & \textbf{Description} \\
    \midrule
    Candidate Ratio ($|\mathcal{C}|/|\mathcal{D}_{train}|$) & $\{0.001, 0.01, 0.1\}$ & The ratios w.r.t. training dataset of the candidate set size \\
    Watermark Ratio ($|\mathcal{W}|/|\mathcal{C}|$) & 1/2 & The ratio of watermarks applied w.r.t. the total candidate number\\
    Duplication Factor ($\textit{d}_{cand=train}$) & 5 & Duplicatation factor for watermarking w.r.t. $\mathcal{D}_{\text{cand}}=\mathcal{D}_{\text{train}}$ \\
   
    Training Split & 0.8 : 0.2 & Training dataset split between $\mathcal{D}_{\text{train}}$ and $\mathcal{D}_{\text{cand}}$ when $\mathcal{D}_{\text{cand}}\neq\mathcal{D}_{\text{train}}$\\
    Test Split & 0.8 : 0.2 & Testing dataset split between $\mathcal{D}_{\text{test}}$ and $\mathcal{D}_{\text{fine}}$\\
    \bottomrule
    \end{tabular}
\end{table*}

\end{document}